\documentclass[a4paper,fleqn,longmktitle]{cas-dc}

\usepackage[numbers]{natbib}
\usepackage[nolist]{acronym}
\newcommand{\etal}{\textit{et~al.\-}}

\def\tsc#1{\csdef{#1}{\textsc{\lowercase{#1}}\xspace}}
\tsc{WGM}
\tsc{QE}
\tsc{EP}
\tsc{PMS}
\tsc{BEC}
\tsc{DE}
\tsc{TW}
\begin{document}
\let\WriteBookmarks\relax
\def\floatpagepagefraction{1}
\def\textpagefraction{.001}
\newcommand{\cmark}{\ding{51}}%
\newcommand{\xmark}{\ding{55}}%
\shorttitle{The MIDOG 2025 Challenge}
\shortauthors{M. Aubreville et~al.}

\title [mode = title]{Mitosis Detection in the Wild: Multi-Tumor and Context-Aware Generalization in the MIDOG 2025 Challenge}                      
%\tnotemark[1,2]

%\tnotetext[2]{The second title footnote which is a longer text matter
%   to fill through the whole text width and overflow into
%   another line in the footnotes area of the first page.}

%%%%%%%%%%%%%%%%%%%%%%%%%%%%%%%%%%%%%%%%%%%%%%%%%%%%%%%%%%
%%%%%      THE MAIN TECHNICAL CREW                   %%%%%
%%%%%%%%%%%%%%%%%%%%%%%%%%%%%%%%%%%%%%%%%%%%%%%%%%%%%%%%%%

\author[1]{Marc Aubreville}

%\tnotetext[1]{}
% Corresponding author indication
%\cormark[1]
%\fnmark[1]
\ead{marc.aubreville@hs-flensburg.de}
%\ead[url]{deepmicroscopy.org}

\credit{Conceptualization,  Data curation, Methodology, Writing -- original draft, Writing -- review \& editing, Project Administration, Funding acquisition}

\affiliation[1]{organization={Flensburg University of Applied Sciences},
                city={Flensburg},
                country={Germany}}
                
\author[2]{Jonas Ammeling}
\affiliation[2]{organization={Technische Hochschule Ingolstadt},
                city={Ingolstadt},
                country={Germany}}
\credit{Conceptualization, Data curation, Methodology, Writing -- review \& editing}

\author[1]{Sweta Banerjee}
\credit{Conceptualization, Data curation, Methodology, Writing -- review \& editing}

%%%%%%%%%%%%%%%%%%%%%%%%%%%%%%%%%%%%%%%%%%%%%%%%%%%%%%%%%%
%%%%%        THE PATHOLOGY EXPERTS                   %%%%%
%%%%%%%%%%%%%%%%%%%%%%%%%%%%%%%%%%%%%%%%%%%%%%%%%%%%%%%%%%

\author[3]{Viktoria Weiss}
\credit{Data curation, Writing -- review \& editing}
\affiliation[3]{organization={Pathology Unit, University of Veterinary Medicine},
                city={Vienna},
                country={Austria}}

\author[4]{Taryn A. {Donovan}}
\affiliation[4]{organization={Schwarzman Animal Medical Center},
                city={New York},
                country={USA}}
\credit{Data curation, Writing -- review \& editing}

\author[5]{Robert Klopfleisch}
\affiliation[5]{organization={Institute of Veterinary Pathology, Freie Universität Berlin},
                city={Berlin},
                country={Germany}}
\credit{Data curation, Writing -- review \& editing}

%%%%%%%%%%%%%%%%%%%%%%%%%%%%%%%%%%%%%%%%%%%%%%%%%%%%%%%%%%
%%%%%          PARTICIPANTS (RANK ORDER)             %%%%%
%%%%%%%%%%%%%%%%%%%%%%%%%%%%%%%%%%%%%%%%%%%%%%%%%%%%%%%%%%

% Team  wildsquirrel
\author[6]{Jiaqi Lv}

\credit{Methodology}
\affiliation[6]{organization=VISION Lab, Tissue Image Analytics Centre, Department of Computer Science, University of Warwick, city= Coventry, country= UK}

\author[6]{Shan E Ahmed Raza}

\credit{Methodology}
% Team  RaphaelBourgade
\author[7]{Raphaël Bourgade} % I removed ,34,35,36 --> strictly only one per person

\credit{Methodology}
\affiliation[7]{organization=Centre for Computational Biology, MINES Paris - PSL University, city= Paris, country= France}
% Pleae, strictly only one affiliation per person. 
%\affiliation[34]{organization=Institut Curie - U1331 Computational Oncology, city= Paris, country= France}
%\affiliation[35]{organization=INSERM U1331, city= Paris, country= France}
%\affiliation[36]{organization=Nantes University Hospital, city= Nantes, country= France}

\author[7]{Thomas Walter} % I removed ,34,35, strictly only one per person
\credit{Methodology, Writing -- review \& editing}

% Team  ytopuz53
\author[8]{Yasemin Topuz}

\credit{Methodology}
\affiliation[8]{organization=Vision Research Group, Department of Computer Engineering, Yildiz Technical University, city= Istanbul, country= Türkiye.}

\author[8]{Songül Varlı}

\credit{Methodology}
% Team  cacfek
\author[9]{Charles-Antoine Collins-Fekete}

\credit{Methodology,Writing -- review \& editing}
\affiliation[9]{organization=Department of Medical Physics and Biomedical Engineering, University College London, city= London, country= UK}

\author[9]{Zhuoyan Shen}

\credit{Methodology}
% Team  navyasri.kelam 
\author[10]{Navya Sri Kelam}

\credit{Methodology}
\affiliation[10]{organization=AIRA MATRIX Private Limited, city= Mumbai, country= India}

\author[10]{Nitin Singhal}

\credit{Methodology}
% Team  christian.marzahl
\author[11]{Christian Marzahl}

\credit{Methodology}
\affiliation[11]{organization=Gestalt Diagnostics, city= Spokane WA 99202, country= USA}

\author[11]{Brian Napora}

\credit{Methodology}
% Team  tengyoux
\author[12]{Tengyou Xu}

\credit{Methodology}
\affiliation[12]{organization=Department of Electrical and Computer Engineering, University of California, Los Angeles, city= Los Angeles, country= USA}

\author[13]{Hongyan Gu}
\affiliation[13]{organization=Department of Pathology and Laboratory Medicine, University of Kansas Medical Center, city=Kansas City, country=USA}

\credit{Methodology}
% Team  masarno
\author[14]{Mario Vento}

\credit{Methodology}%
\affiliation[14]{organization=Department of Information and Electrical Engineering and Applied Mathematics (DIEM) University of Salerno Fisciano, city=Salerno, country= Italy}

\author[14]{Gennaro Percannella}

\credit{Methodology}
% Team  piotrgiedziun
\author[15]{Norbert Ropiak} % email unknown - contacted via Piotr

\credit{Methodology}
\affiliation[15]{organization=Cancer Center Sp. z o. o., city= Wroclaw, country= Poland}

\author[16]{Izabela Wasiak} % email unknown - contacted via Piotr
\affiliation[16]{Pathology Department, 10th Military Research Hospital in Bydgoszcz, city = Bydgoszcz, country = Poland}

\credit{Methodology}
% Team  SZTU-134
\author[17]{Jie Xiao}

\credit{Methodology}
\affiliation[17]{organization={College of Health Science and Environmental Engineering, Shenzhen Technology University}, city=Shenzhen, country= China}

\author[17]{Shaojun Liu} % email unknown - contacted via Jie Xiao

\credit{Methodology}
% Team  schoe
\author[18]{Seungho Choe}

\credit{Methodology}
\affiliation[18]{organization=Image Analysis in Medicine Lab (IAMLAB), Electrical, Computer and Biomedical Engineering, Toronto Metropolitan University, Toronto, city= ON, country= Canada}

\author[18]{April Khademi}

\credit{Methodology}
% Team  vidushiwalia
\author[19]{Vidushi Walia}

\credit{Methodology}
\affiliation[19]{organization=TCS Research, Tata Consultancy Services Ltd., city= Hyderabad, country= India}

\author[19]{Sujatha Kotte}

\credit{Methodology}
% Team  npic-ab
\author[20]{Andrew Broad}

\credit{Methodology}
\affiliation[20]{organization=National Pathology Imaging Co-operative, city= Leeds Teaching Hospitals NHS Trust, country= UK}

\author[20]{Alex Wright} 

\credit{Methodology}
% Team  guillaume.balezo
\author[7]{Guillaume Balezo} % Sorry, strictly only one affiliation per person. I removed 31
%\affiliation[31]{organization=Digital R\&D Sanofi, city= Paris, country= France}
\credit{Methodology}

%\author[7]{Hana Feki} -> Removed, as per mail from Thomas Walter, June 4th, 22:44
%\credit{Methodology}

% Team  nasires
\author[6]{Esha Sadia Nasir}

\credit{Methodology}
\author[6]{Mostafa Jahanifar}

\credit{Methodology}
% Team  yohsuke.yamagishi
\author[21]{Yosuke Yamagishi}

\credit{Methodology}
\affiliation[21]{organization=Division of Radiology and Biomedical Engineering, Graduate School of Medicine, The University of Tokyo, city= Tokyo, country= Japan}

\author[21]{Shouhei Hanaoka} % no email, contacted via Yosuke

\credit{Methodology}
% Team  masarno
% Author  Mattia Sarno  already found, not duplicating. 
\author[14]{Mattia Sarno}
\author[14]{Francesco Tortorella} 

\credit{Methodology}
% Team  zerostarcraft
\author[22]{Biwen Meng}

\credit{Methodology}
\affiliation[22]{organization=School of AI and Advanced Computing, Xi'an Jiaotong-Liverpool University, city= Suzhou, country= China}

\author[22]{Jingxin Liu}

\credit{Methodology}
% Team  krausara
\author[23]{Sara Krauss}

\credit{Methodology}
\affiliation[23]{organization=IT-Infrastructure for Translational Medical Research, Faculty of Applied Computer Science, city= University of Augsburg, country= Germany}

\author[24]{Daniel Hieber}
\affiliation[24]{organization=Institute of Neuropathology, Ulm University Medical Center, Faculty of Medicine, Ulm University, city=Ulm, country=Germany}

\credit{Methodology}
% Team  lrc9859
\author[10]{Lavish Ramchandani}

\credit{Methodology}

\author[10]{Dev Kumar Das} % no known email - contacted via Lavish

\credit{Methodology}
% Team  be_yuan
\author[25]{Mieko Ochi}

\credit{Methodology}
\affiliation[25]{organization=Department of Pathology, Japanese Red Cross Medical Center, city= Tokyo, country= Japan}

\author[25]{Yuan Bae}

\credit{Methodology}
% Team  piotrgiedziun
\author[26]{Piotr Giedziun}

\credit{Methodology}
\affiliation[26]{organization=Department of Artificial Intelligence, Wroclaw University of Science and Technology, city= Wrocław, country= Poland}

\author[15]{Mateusz Maniewski} % no email, contacted via Piotr
\credit{Methodology}
% Team  cacfek
% Author  Charles-Antoine Collins-Fekete  already found, not duplicating. 

% Author  Zhuoyan Shen  already found, not duplicating. 

% Team  schoe
% Author  Seungho Choe  already found, not duplicating. 

% Author  April Khademi  already found, not duplicating. 

% Team  saipradeepvg
\author[18]{Vangala Govindakrishnan Saipradeep}

\credit{Methodology}
\author[18]{Naveen Sivadasan}

\credit{Methodology}
% Team  navyasri.kelam
% Author  Navya Sri Kelam  already found, not duplicating. 

% Author  Nitin Singhal  already found, not duplicating. 

% Team  Leire
\author[27]{Leire Benito-Del-Valle}

\credit{Methodology}
\affiliation[27]{organization=TECNALIA, Basque Research and Technology Alliance (BRTA), Parque Tecnológico de Bizkaia, city= C/ Geldo. Edificio 700, country= E-48160 Derio- Bizkaia (Spain)}

\author[27]{Adrian Galdran}

\credit{Methodology}
% Team  Kaustubh_Atey
\author[28]{Kaustubh Atey}

\credit{Methodology}
\affiliation[28]{organization=Centre for Machine Intelligence and Data Science, city= Indian Institute of Technology Bombay, country= India}

\author[28]{Sameer Anand Jha} % no email, contacted via Kaustubh

\credit{Methodology}
% Team  mirazzak
\author[29]{Adinath Dukre}

\credit{Methodology}
\affiliation[29]{organization=MBZUAI, city= Abu Dhabi, country= UAE}

\author[29]{Imran Razzak}

\credit{Methodology}
% Team  mlafarge
\author[30]{Maxime W. Lafarge}

\credit{Methodology}
\affiliation[30]{organization=Computational and Translational Pathology Lab, Department of Biomedical Engineering, University of Basel, city= Allschwil, country= Switzerland}

\author[30]{Viktor H. Koelzer}

\credit{Methodology, Writing -- review \& editing}
% Team  chillice
% Author  Tengyou Xu  already found, not duplicating. 

%\author[12]{chillice-1}

%%%%%%%%%%%%%%%%%%%%%%%%%%%%%%%%%%%%%%%%%%%%%%%%%%%%%%%%%%
%%%%%   THE REMAINING ORGANIZATION TEAM              %%%%%
%%%%%%%%%%%%%%%%%%%%%%%%%%%%%%%%%%%%%%%%%%%%%%%%%%%%%%%%%%
\author[1]{Nils Porsche}
\credit{Visualization, Writing -- review \& editing}

\author[31]{Nikolas Stathonikos}
\affiliation[31]{organization={University Medical Center Utrecht},
				 city={Utrecht},
				 country={The Netherlands}}

\credit{Conceptualization,  Data curation, Methodology, Writing -- review \& editing, Project Administration}

\author[32]{Mitko Veta}
\affiliation[32]{organization={TU Eindhoven},
                city={Eindhoven},
                country={The Netherlands}}
\credit{Conceptualization,  Writing -- review \& editing, Project Administration}
                
\author[33]{Dominik Hirling}
\credit{Conceptualization,  Writing -- review \& editing, Data curation, Project Administration}
\affiliation[33]{organization={HUN-REN Biological Research Centre, Institute of Biochemistry},
				 city={Szeged},
				 country={Hungary}}  

\author[33]{Zsanett Zsófia Iván}
\credit{Data curation}

\author[33]{Peter Horvath}

\credit{Conceptualization,  Data curation, Writing -- review \& editing, Project Administration, Funding acquisition}

\author[34]{Katharina Breininger}

\affiliation[34]{organization={Julius-Maximilians-Universität Würzburg},
                city={Würzburg},
                country={Germany}}
\credit{Conceptualization,  Writing -- review \& editing, Project Administration}

\author[3]{Christof A. {Bertram}}
\ead{christof.bertram@vetmeduni.ac.at}
\credit{Conceptualization,  Data curation, Writing -- original draft, Writing -- review \& editing, Project Administration, Funding acquisition}

%\cortext[cor1]{Corresponding author}
%\cortext[cor2]{Principal corresponding author}
%\fntext[fn1]{M.A. and S.B. acknowledge support from the DFG ... etc...}
%\fntext[fn2]{Another author footnote, this is a very long footnote and
%  it should be a really long footnote. But this footnote is not yet
%  sufficiently long enough to make two lines of footnote text.}

%\nonumnote{This note has no numbers. In this work we demonstrate $a_b$
%  the formation Y\_1 of a new type of polariton on the interface
%  between a cuprous oxide slab and a polystyrene micro-sphere placed
%  on the slab.
%  }

% Then just before \begin{abstract}:
\begin{abstract}
Automated mitotic figure detection is a well-established task in computational pathology. Previous benchmarks addressing robustness have focused on scanner-induced domain shifts; however, real-world clinical use requires models that are robust to the much broader variability present in the histological landscape. The third edition of the MItosis DOmain Generalization (MIDOG) challenge, held in 2025, was designed to evaluate algorithmic performance across unprecedented biological and contextual diversity. The challenge was composed of two tracks: 1) mitotic figure object detection, and 2) classification of mitotic figures into normal and atypical morphologies. 
%Methods:
We curated a comprehensive test dataset of 365 cases, encompassing 12 distinct human, canine and feline tumor types, digitized across multiple scanning platforms. Moving beyond traditional hand-selected hotspots, this challenge additionally required detection in random tissue areas (representative of the whole slide image) and challenging areas (areas rich in imposters). Participants were tasked with developing architectures capable of maintaining high precision despite these varying levels of difficulty and tissue architecture.
%Results:
For track 1, predictions were submitted by 18 teams, with $F_1$ scores of up to 0.740. In the second track, there were 21 submissions with balanced accuracy of up to 0.908. Our analysis reveals that while most models perform reliably in traditional hotspots, significant performance degradation occurs in challenging regions, where the false positive rate increased by 208\%. Furthermore, performance varied significantly across the 12 tumor types, highlighting "blind spots" in current state-of-the-art architectures when encountering rare or highly pleomorphic cancer types. 
We evaluated the effectiveness of ensembling and test-time augmentation (TTA) and found the former to have a consistently positive effect (mean increase of 1.5 percentage points in $F_1$ score in track 1 and 1.3 in balanced accuracy in track 2), whereas TTA showed no relevant improvement.
%Conclusions:
The MIDOG 2025 challenge demonstrates that "in the wild" mitosis detection remains a significant hurdle. The transition from hotspot-only evaluation to a multi-contextual framework (hotspot, challenging, and random regions) provides a more realistic proxy for clinical reliability. 
\end{abstract}

\begin{graphicalabstract}
\includegraphics[width=\textwidth]{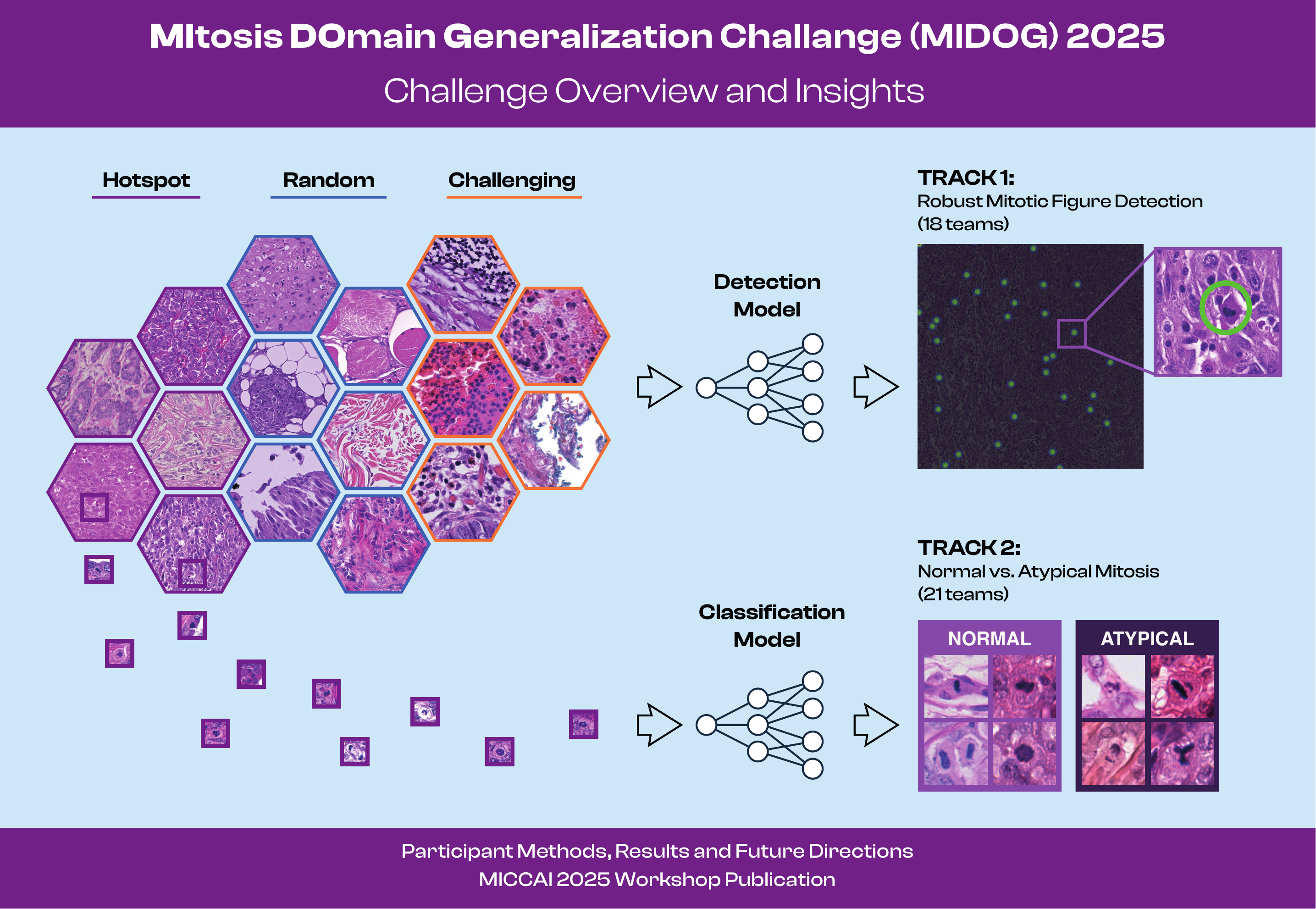}
\end{graphicalabstract}

\begin{highlights}
\item Performance of automatic mitosis detection models collapses outside curated hotspot regions. Evaluating in random and challenging (rich in imposters) regions more than tripled the false positive rate (increase of 208\%), exposing a clinical reliability gap.
\item Biological diversity reveals blind spots. Across 365 cases and 12 human, canine, and feline tumor types, top teams hit 0.740 F1 (detection) and 0.908 balanced accuracy (atypical classification) — but mitosis detection consistently underperformed on rare and highly pleomorphic tumors.
\item Ensembling helps, TTA does not. Model ensembling gave consistent gains (average increase of 1.5 percentage points in overall F1, 1.3 percentage points in overall balanced accuracy); test-time augmentation produced no meaningful improvement.
\end{highlights}

\begin{keywords}
pattern recognition challenge \sep mitosis detection \sep atypical mitosis \sep domain generalization
\end{keywords}

\begin{acronym}
% ── Staining & histology ──────────────────────────────────────────────────────
\acro{HE}[H\&E]{hematoxylin and eosin}
\acro{PHH3}[PHH3]{phosphohistone H3}
\acro{IHC}[IHC]{immunohistochemistry}

% ── Imaging & scanning ────────────────────────────────────────────────────────
\acro{ROI}[ROI]{region of interest}
\acro{WSI}[WSI]{whole slide image}
\acro{MPP}[MPP]{microns per pixel}

% ── Biological entities ───────────────────────────────────────────────────────
\acro{MF}[MF]{mitotic figure}
\acro{AMF}[AMF]{atypical mitotic figure}
\acro{NMF}[NMF]{normal mitotic figure}

% ── Challenge & conference names ──────────────────────────────────────────────
\acro{MIDOG}[MIDOG]{Mitosis Domain Generalization}
\acro{MICCAI}[MICCAI]{International Conference on Medical Image Computing and Computer Assisted Intervention}
\acro{AMIDA}[AMIDA]{Automatic Mitosis Detection Algorithm}
\acro{TUPAC}[TUPAC]{Tumor Proliferation Assessment Challenge}

% ── Metrics ───────────────────────────────────────────────────────────────────
\acro{FROC}[FROC]{free-response receiver operating characteristic}
\acro{ROC}[ROC]{receiver operating characteristic}
\acro{AUC}[AUC]{area under the curve}
\acro{AP}[AP]{average precision}
\acro{FPPI}[FPPI]{false positives per image}
\acro{TP}[TP]{true positive}
\acro{FP}[FP]{false positive}
\acro{FN}[FN]{false negative}
\acro{BA}[BA]{balanced accuracy}

% ── Statistical concepts ──────────────────────────────────────────────────────
\acro{ICC}[ICC]{intraclass correlation coefficient}
\acro{CI}[CI]{confidence interval}

% ── Machine learning techniques ───────────────────────────────────────────────
\acro{TTA}[TTA]{test-time augmentation}
\acro{EMA}[EMA]{exponential moving average}
\acro{LoRA}[LoRA]{low-rank adaptation}

% ── Model architectures & components ─────────────────────────────────────────
\acro{CNN}[CNN]{convolutional neural network}
\acro{FCOS}[FCOS]{fully convolutional one-stage detector}
\acro{DETR}[DETR]{detection transformer}
\acro{YOLO}[YOLO]{you only look once}
\acro{VRAM}[VRAM]{video random access memory}

% ── Clinical & pathology terms ────────────────────────────────────────────────
\acro{MC}[MC]{mitotic count}
\acro{HPF}[HPF]{high-power field}

% ── Institutions & databases ──────────────────────────────────────────────────
\acro{DA}[DA]{digital archive}
\acro{VMU}[VMU]{Vienna Medical University}
\acro{FUB}[FUB]{Freie Universit\"at Berlin}
\acro{UMC}[UMC]{University Medical Center}
\acro{AJCC}[AJCC]{American Joint Committee on Cancer}
\acro{WHO}[WHO]{World Health Organization}

% ── Software & tools ─────────────────────────────────────────────────────────
\acro{EXACT}[EXACT]{EXpert Annotation Collaboration Tool}
\end{acronym}

\maketitle

\section{Introduction}
Quantification of cells that undergo division, represented by mitotic figures in histologic images, is an important task for assessing cancer aggressiveness. As part of routine pathologic evaluation of many tumor types, the number of \acp{MF} (i.e., the \ac{MC}) in a predefined region of the tumor (often defined as 2 to 2.37 $mm^2$) with the highest density (hotspot) are manually counted~\citep{fitzgibbons2023protocol,mcniel1997evaluation,kiupel2011proposal,bertram2024mitoticC,bertram2024mitoticF,who2022endocrine}. However, this task has a well-known inter-rater disagreement for classifying \acp{MF} against other histologic structures~\cite{malon2012mitotic,veta2016mitosis} as well as a sampling bias for selection of the tumor region~\cite{aubreville2020deep}, which can impact therapeutic decisions. This measurement variability may be reduced by the use of automatic mitosis detection tools in the diagnostic workflow. 

Consequently, the mitosis recognition task is well-esta\-blished in computational pathology. Starting with the MITOS challenge in 2012~\cite{LUDOVIC20138}, the community has conducted several additional challenges in the subsequent years (AMIDA 13~\cite{veta2015assessment}, MITOS-ATYPIA 2014~\cite{roux2014mitos}, TUPAC16 \cite{veta2019predicting}, MIDOG 2021 ~\cite{aubreville2023mitosis}, and MIDOG 2022 ~\cite{aubreville2024domain}), fostering the validation of modern pattern recognition architectures for this task. 
While early challenges focused on breast cancer and offered limited domain diversity, the two \ac{MIDOG} challenges specifically targeted the most important property of clinically applicable pattern recognition solutions: generalization to diverse and previously unseen sample distributions. In the first (2021) MIDOG challenge edition, images from various whole slide scanners were introduced, testing the generalization to unseen whole-slide image scanners~\cite{aubreville2023mitosis}. In the following year's challenge the scope widened to generalization across different tumor types, introducing a test set of 100 cases across ten different (and partially unseen) tumors from human, canine and feline specimens~\cite{aubreville2023mitosis}. The limitation of all previous challenges is that the detection task was not performed on entire \acp{WSI}, but on expert-selected hotspot regions within the tumor, where the highest \ac{MC} was assumed based on a high cellular density and swift screening for \acp{MF}, while excluding challenging areas with an increased occurrence of structures with morphologic overlap to \acp{MF} (i.e., imposters). 

Restricting datasets to hotspot regions limits the applicability of derived models to routine pathology workflows. As shown in prior research~\citep{aubreville2020deep,bertram2020computerized}, a major source of inter-pathologist variability in the \ac{MC} is the localization of the most proliferative \ac{ROI}, which introduces sampling bias. Consequently, in traditional \ac{MC} assessment, hotspots (believed to be most indicative of biological tumor behavior) are often not identified ~\cite{aubreville2020deep}, which can negatively affect accurate tumor prognostication and treatment decision. This provides strong motivation for the computerized detection of \acp{MF} on entire \acp{WSI}, and potentially also across multiple sections of the same tumor~\citep{stathonikos2024breast,bertram2022computer}. Consequently, these completely automatic workflows are fundamental for a meaningful deployment in a clinical routine. This necessitates validating model performance on whole slide images (or representative regions thereof) to investigate if a reliable application of models to this use case scenario is possible.

Motivated by these considerations, the \ac{MIDOG} 2025 challenge is the first large-scale algorithmic evaluation that extends the scope beyond hotspot regions, evaluating also on \acp{ROI} representing all the other regions of the \ac{WSI}, including particularly challenging \acp{ROI}, assumed to be rich in imposters (i.e., structures that are suspected to lead to false positive predictions). 

Besides detecting \acp{MF} across \acp{WSI}, another task of interest is the classification of \acp{MF} into normal morphologies and \acp{AMF}. An \ac{AMF} can be histologically identified in cells exhibiting abnormal division due to chromosome segregation errors~\citep{donovan2021mitotic,melba:2026:006:banerjee}. This asymmetric cell division can lead to daughter cells with an abnormal number of chromosomes (aneuploidy), a relevant mechanism how tumor cells can accumulate mutations required for cancer progression \citep{gisselsson2008classification,sdeor2024aneuploidy}. Aneuploidy leads to copy number alterations of numerous genes at once, and thereby can have a massive effect on cell behavior and function, overall leading to a more aggressive behavior of the tumor. There is growing evidence linking a high frequency and proportion of \acp{AMF} to reduced patient survival in breast cancer~\citep{ohashi2018prognostic,lashen2022characteristics} and other cancer types~\citep{jin2007distinct,kalatova2015tripolar,bertram2023atypical,matsuda2016mitotic}. A comprehensive, large-scale study by Jahanifar et al.~\cite{jahanifar2025pan} has recently confirmed the prognostic relevance of \acp{AMF} rates across a diverse spectrum of tumor types, thereby underscoring the need for further research into this prognostic marker. Motivated by these findings, some recent studies have investigated computerized classification approaches~\citep{banerjee2026benchmarking,bertram2025histologic}. These studies propose connecting the \ac{AMF} model as a subsequent stage to the \ac{MF} model, making \acp{AMF} a compelling choice for a second track in the MIDOG 2025 challenge.

\begin{figure*}
    \includegraphics[width=\textwidth]{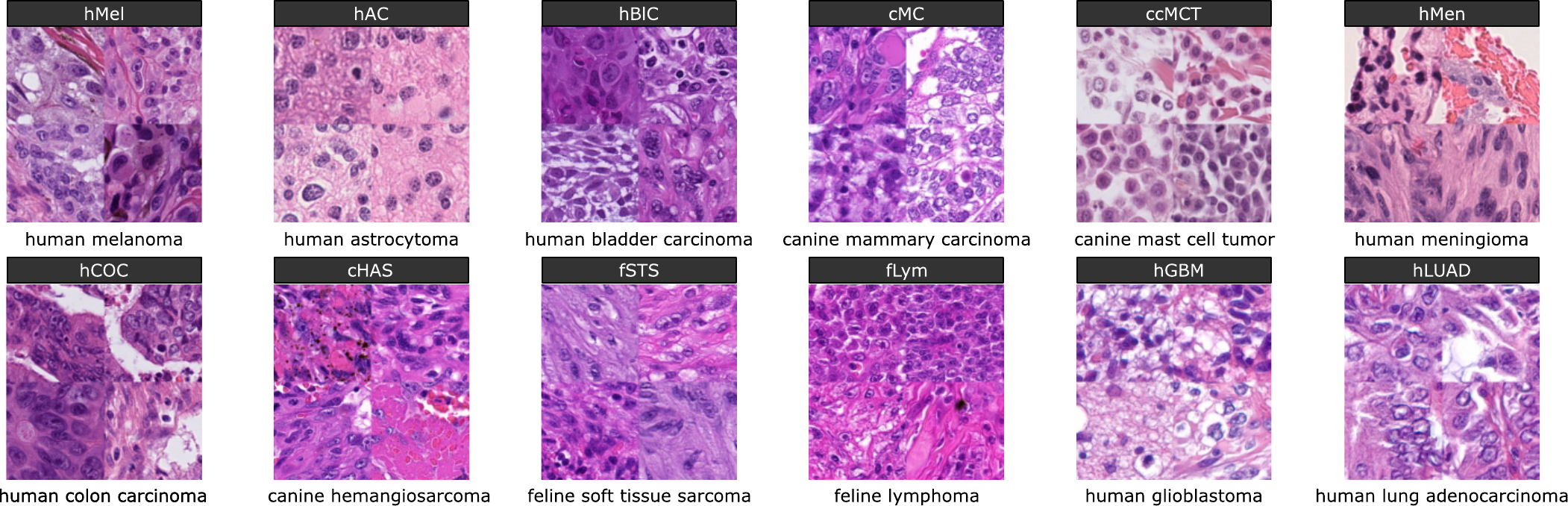}
    \caption{Domains of the test set of the MIDOG 2025 challenge. For each domain, both the abbreviation chosen as well as the full name is given. Shown are four random samples from within the hotspot regions of interest.}
\end{figure*}
\subsection*{Challenge format and task}

The \ac{MIDOG} 2025 challenge was organized as a satellite event of the \ac{MICCAI} 2025, following an open call for participation and peer review of the challenge design~\cite{ammeling_2025_15077361}. Information about the challenge was made available through the challenge website\footnote{https://midog2025.deepmicroscopy.org}. The challenge comprised two independent tracks, with participants able to submit to either or both.

\paragraph{Track 1: Mitotic figure detection.}
This track required the localization of all \acp{MF} within a given \ac{ROI}, across a broad range of tumor types and digitization devices. In contrast to all prior editions of the \ac{MIDOG} challenge and other mitosis detection challenges, evaluation was not restricted to expert-selected hotspot \acp{ROI} but extended to randomly sampled tissue regions and expert-selected challenging regions rich in imposters. Specifically, for each of the 122 test cases (twelve domains, at least ten per domain), at least three \acp{ROI} of 2~mm$^2$ each were extracted following the three sampling strategies described below, yielding 365 \acp{ROI} in total.
Each case corresponded to an individual patient.

\begin{figure}
    \centering
    \includegraphics[width=1.0\linewidth]{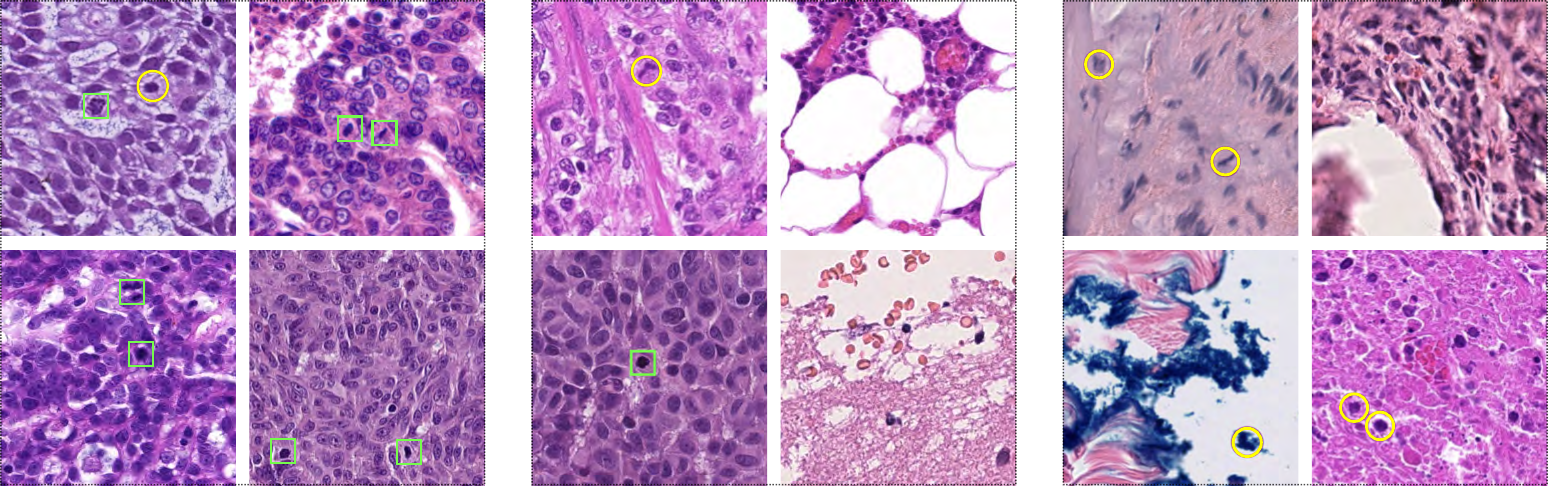}
    \caption{Region types in track 1 of the MIDOG 2025 challenge. Left panel shows hotspot regions, middle panel shows random regions and right panel shows challenging regions. Green squares indicate PHH3-confirmed mitotic figures, yellow circles indicate false detections by our reference model.}
    \label{fig:region_types}
\end{figure}

\paragraph{ROI sampling strategies.}
Three complementary \ac{ROI} types were defined for Track~1 evaluation (see Figure \ref{fig:region_types}):

\begin{itemize}
    \item \textbf{Hotspot \acp{ROI}}: regions of high \ac{MF} density within the tumor, selected by a board-certified veterinary pathologist following standard clinical practice. At low magnification, several eligible tumor regions were selected based on high cellular density and optimal  image and tissue quality (e.g., absence of tissue and scan artifacts, necrosis, or marked inflammation, and avoiding areas with delayed fixation). Subsequently these regions were screened at higher magnification and the regions with the suspected highest \ac{MF} density were selected. 
    \item \textbf{Random \acp{ROI}}: regions sampled uniformly at random from the tissue area of the \ac{WSI} (following a basic thresholding algorithm), subject to a minimum tissue coverage threshold of 80\%. Random \acp{ROI} are statistically representative of the full tissue composition of the slide and serve as the primary proxy for realistic whole-slide evaluation conditions.
    \item \textbf{Challenging \acp{ROI}}: regions deliberately selected by a board-certified veterinary pathologist for their high density of \ac{MF} imposters -- morphological structures that may be mistaken by the models for \acp{MF}, including apoptotic and necrotic cells, hyperchromatic nuclei, inflammatory cells, and ink for margin marking. Additionally, \acp{ROI} with scan artifacts (e.g., out-of-focus regions) and sectioning artifacts (e.g., tissue folds) were included. These regions represent a curated worst case for false positive generation.
\end{itemize}

\paragraph{Track 2: Atypical mitotic figure classification.}
The second task required the binary classification of image patches sized $128\times 128$ px, centered on confirmed \acp{MF} as either normal or atypical. This track was motivated by the emerging prognostic relevance of \acp{AMF} and provided the first large-scale, multi-domain, multi-scanner dataset of normal and atypical mitotic figures for public use, comprising 11,939 \acp{MF} across the seven domains~\cite{weiss_2025_16044804} provided by the MIDOG++~\cite{aubreville2023comprehensive} dataset.

\paragraph{Submission format and timeline.}
% Disqualification memo:
% Carrot: Kicked out of Grand-Challenge (fair use policy)
% proproscrinator31: No preprint received
% gdeotale123: No preprint received
% rishadanand: Preprint not publicly available

MIDOG 2025 was a one time event with fixed submission deadline for which 164 persons registered on the grand-challenge platform. For both tracks, participants were required to submit self-contained Docker containers, which included their complete inference pipeline, ensuring reproducibility and enabling blind evaluation on the withheld test set. Example implementations, including baseline algorithms and instructions for submission, were made available on GitHub\footnote{Track 1: \url{https://github.com/DeepMicroscopy/MIDOG25_T1_reference_docker} and Track 2: \url{https://github.com/DeepMicroscopy/MIDOG25_T2_reference_docker}}. The baseline method for Track 1 was provided as FCOS\cite{tian2019fcos} detection model and for Track 2 we provided an EfficientNetV2 \cite{tan2021efficientnetv2}-based classifier \cite{proc_Banerjee2026}. The evaluation methods (not including the ground truth of the test set) were provided on GitHub as well.\footnote{Track 1: \url{https://github.com/DeepMicroscopy/MIDOG25_T1_evaluation_docker}, Track 2: \url{ https://github.com/DeepMicroscopy/MIDOG25_T2_evaluation_docker}}

A preliminary test set of 20 cases across four tumor domains was made available for automated evaluation by the grand-challenge platform from August~15 2025 to allow validation of the submission pipeline. Submissions of 31 and 35 teams on the preliminary set for Track~1 and Track~2, respectively, were received. The final test set submission window opened on August~30 and closed on September~1, 2025 and teams were allowed to submit once to it. Final submissions were accepted from 18 teams in Track~1 and 21 teams in Track~2. From these, 3 teams were disqualified for failing to supply a preprint or for violating the terms of Track 1. 

\paragraph{Participation rules.}
The organizers provided training data for both tracks, released under Creative Commons licenses. The use of additional data sets was permitted, given that these were publicly available without conditions to all participants. Access to the test set and test set labels was only provided to the organizers M.A., J. A. and S.B.

Researchers belonging to the institutes of the organizers were not allowed to participate to avoid potential conflict of interest. Participants were permitted to publish papers including their official performance on the challenge data set, given proper reference of the challenge, without embargo time. Participants were asked to publish a concise description of their method and results on a preprint server or on a general-purpose open-access repository. Public release of the source code was not mandatory. The first three positions in each track were awarded with monetary prizes. Results were announced at the conference workshop.

\paragraph{Peer review.}
Each submitted preprint was reviewed independently by at least two expert reviewers, who assessed contributions according to three criteria: technical clarity (whether the approach and training procedure were clearly described), innovation (whether the approach was novel or introduced an innovative adaptation of an existing method), and formal quality (whether the paper was well-structured and clearly written). Of the 31 preprints submitted alongside final test set entries, 27 were accepted for presentation at the \ac{MIDOG} 2025 workshop, held in conjunction with \ac{MICCAI} on September~23, 2025. Teams whose papers achieved the highest peer review scores, as well as the top-performing teams on the final leaderboard, were invited to contribute extended versions of their papers to the proceedings, which underwent a further independent full peer review.

%In the most recent third edition of MIDOG, this generalization was further extended, with the test set now comprising 365 histopathology images spanning each $2mm^2$ from twelve different unseen tumor types. These were, however, not only selected from the region of highest cellular density of the tumor, as done in all previous challenges on mitosis detection, but extend to a contextual framework that encompasses three distinct modes. 

%For one, and representing the traditional use case of regions selected for the mitotic count by an expert,  traditional hotspot \acp{ROI} formed one third of the test set. Moreover, representing a use case where algorithms are applied to the whole slide image, we included random regions selected from the tissue area of the slide. 
% Mitosis Detection is clinically relevant for tumor grading

%  

\section{Material and Evaluation Methods}

%We utilized the same 100 \acp{WSI} that were used to sample the multi-domain test set of the predecessor \ac{MIDOG} 2022 challenge~\cite{aubreville2024domain}, and additionally added two new domains (human glioblastoma and human lung adenocarcinoma), yielding 122 \acp{WSI} in total across twelve domains. 

\subsection{Material}
Training sets: For the task of mitosis detection, a good variety of datasets already existed prior to the challenge (e.g.,~\cite{aubreville2020completely, aubreville2023comprehensive, bertram2019large,bertram2020pathologist, roux2014mitos, veta2019predicting}), including the MIDOG++ dataset~\cite{aubreville2023comprehensive}, which is an extended version of the dataset used in the predecessor challenge in 2022. The organizers therefore decided to not release an additional dataset for MIDOG 2025 for this task. However, given the field of \ac{AMF} detection was still in its infancy, with only the Ami-Br dataset~\cite{bertram2025histologic} being available prior to the challenge, we released an additional public training dataset of \acp{AMF}, encompassing annotations for all 11,939 \acp{MF} of the MIDOG++ dataset~\cite{weiss_2025_16044804}. For the annotation process and dataset characteristics of the training datsets we refer to the previous publications. 

Test sets: For this challenge, we sourced twelve tumor types as the test set, as shown in Table \ref{tab:domains}. The first ten out of those were previously used for the MIDOG 2022 challenge. Per domain, we used ten tumor cases, corresponding to ten patients, from which one representative \ac{WSI} was digitized. In addition to the cases for MIDOG 2022, we sourced two human glioblastoma (hGBM) and human lung adenocarcinoma (hLUAD) from a new challenge organization partner (University of Szeged, Hungary). Human glioblastoma was represented with ten cases/\acp{WSI}, whereas we incorporated twelve cases/\acp{WSI} of human lung adenocarcinoma. Thus, our dataset comprises \acp{WSI} from 122 patients/cases, split across twelve domains.

For each of the 122 \acp{WSI}, we selected three mutually exclusive \acp{ROI}, each one for the sampling category of \textit{hotspot region}, \textit{random region}, and \textit{challenging region} (see Figure~\ref{fig:region_types}). In one case the tissue was too small for our random sampler to yield a suitable, non-overlapping and sufficiently tissue-covered region, yielding a total of 365 \acp{ROI} for the final test set, split across 122 hotspot regions, 121 random regions and 122 challenging regions. The annotation process for these images is outlined below.

Preliminary test set: For validation of the technical function of the participants' docker containers, we  provided access to execution on a four domain, 20 case preliminary test set, as done in previous MIDOG challenges. For this, we reused the preliminary test set of \ac{MIDOG}
2022, encompassing human breast carcinoma, canine osteosarcoma, human lymphoma, and canine pheochromocytoma~\cite{aubreville2024domain}. The organizers notified the participants that this dataset is not a good proxy for the final test set, as it only encompasses hotspot \acp{ROI}, is considerably smaller, and originates from different tumor types, and that hyperparameter tuning on this dataset in the preliminary evaluation phase is discouraged due to this. 

\begin{table*}[htbp]
\centering
\caption{Overview of the domains of the final test set. }
\resizebox{\textwidth}{!}{%
\begin{tabular}{lllll}
\hline
\textbf{Domain} & \textbf{Tumor Type} & \textbf{Tumor Origin} & \textbf{Scanner / Res.}  & \textbf{Origin} \\ \hline
hMel & Human melanoma & Neuroectodermal & Hamamatsu S360; 0.23 $\mu m$/px  &  \multirow{10}{*}{MIDOG 2022~\cite{aubreville2024domain}}\\
hAC & Human astrocytoma & Neuroectodermal & Hamamatsu S60; 0.22 $\mu m$/px &  \\
hBlC & Human bladder carcinoma & Epithelial & 3DHistech P. Scan II; 0.25 $\mu m$/px &  \\
cMC & Canine breast carcinoma & Epithelial & 3DHistech P. Scan II; 0.25 $\mu m$/px & \\
ccMCT & Canine cutaneous mast cell tumor & Mesenchymal & Hamamatsu S360; 0.23 $\mu m$/px &  \\
hMen & Human meningioma & Mesenchymal/Neuroectodermal & Hamamatsu S60; 0.22 $\mu m$/px  & \\
hCoC & Human colon carcinoma & Epithelial & Hamamatsu S360; 0.23 $\mu m$/px &  \\
cHAS & Canine hemangiosarcoma & Mesenchymal & 3DHistech P. Scan II; 0.25 $\mu m$/px &  \\
fSTS & Feline soft tissue sarcoma & Mesenchymal & 3DHistech P. Scan II; 0.25 $\mu m$/px &  \\
fLym & Feline GI lymphoma & Mesenchymal & 3DHistech P. Scan II; 0.25 $\mu m$/px &  \\ \hline
hGBM & Human glioblastoma & Neuroectodermal & 3DHistech P1000, 0.12 $\mu m$/px & \multirow{2}{*}{University of Szeged} \\
hLUAD & Human lung adenocarcinoma & Epithelial & 3DHistech P1000, 0.12 $\mu m$/px  \\ \hline\end{tabular}%
}
\label{tab:domains}
\end{table*}

\paragraph{Ethics approval:} We received approval by the UMC Utrecht's ethics board (TCBio 20-776), the Regional and Institutional Research Ethics Committee
of the University of Szeged (BM/22651-1/2024) and the ethics board of the medical faculty of FAU Erlangen-Nürnberg (AZ 92\_14B, AZ 193\_18B). For animal samples taken from the diagnostic archive, no ethics approval is required.

\subsection{Annotation}
%Annotations for conducted for the test set; for track 1 only for the new images (while using the previous annotations of \ac{MIDOG} 2022 for the previous images) and for track 2 for all images. 
For the test set of track 1, we annotated only the new images (i.e., the random and challenging \acp{ROI} and the two additional tumor types) while using the previous annotations of \ac{MIDOG} 2022 for those respective images. 
For \ac{MF} annotations we used \ac{IHC}-assisted labeling, using the following image preparation steps. For each sample, after successful digitization of the \ac{HE} stained slide and quality control, we removed the cover slip from the slide, washed out the \ac{HE} stain and applied \ac{IHC} with antibodies against \ac{PHH3}. This antibody immunolabels mitotic figures in early pro- to late telophase, facilitates annotation for \acp{MF}, and was successfully used in prior research~\cite{hendzel1997mitosis,bertram2020computerized,tellez2018whole,aubreville2024domain}. However, as shown by Ganz \etal~\cite{ganz2024information}, the immunopositivity of early prophase \acp{MF}, lacking discriminative morphological features for \acp{MF}, can lead to an information mismatch between the \ac{HE} image and \ac{PHH3}-labeled images. This has been shown to cause annotation bias, requiring an adaptation of purely IHC-based annotation workflows to avoid annotations of morphologically indistinctive cells \cite{ganz2024information}. 

Our \ac{MF} annotation pipeline was composed of two annotation phases. The first phase was conducted by a board-certified veterinary pathologist (CAB) that used \ac{PHH3} images as annotation support. For this, we registered \ac{HE} and \ac{PHH3} \acp{WSI} using manual selection of four corresponding points in both images. We then calculated the affine transform between both using a least squares optimizer. This workflow was implemented by the challenge organizers in the open source EXACT annotation software~\cite{marzahl2021exact}, allowing for the annotating pathology to blend between the \ac{HE} and \ac{PHH3} images using a slider and keyboard shortcuts. In this first annotation step, the expert annotated cells according to five categories:
\begin{enumerate}
    \item PHH3-positive, morphology in \ac{HE} clearly distinctive of \acp{MF}
    \item PHH3-positive, morphology in \ac{HE} suggestive of a \ac{MF}
    \item PHH3-negative, morphology in \ac{HE} suggestive of a \ac{MF}
    \item PHH3-positive, morphology in \ac{HE} suggestive of an imposter
    \item PHH3-negative and morphology resembling, but distinct from a \ac{MF}
\end{enumerate}
%For the second annotations phase, the first two categories were attributed the label of \ac{MF}.  
Annotations of the first three label classes were considered as \acp{MF} and the last two label classes were considered as hard negatives. 
For the second annotation phase, patches with an approximate size of 30 microns were cropped around those annotations from the \ac{HE} image, and blindly shown to a secondary board-certified pathologist (RK), tasked with classifying them into \ac{MF} or non-\ac{MF} imposter cells. The inclusion of hard negatives in the second step ensured no assumptions about the distribution/priors were possible by the second expert. In case of agreement, the annotation was included into the final dataset. In case of disagreement, a third expert (TAD, also a board-certified pathologist) was asked to act as a tie breaker, rendering the final label for the annotation. For the 100 hotspot \acp{ROI} sourced from the previous \ac{MIDOG} 2022 challenge, we reused the \ac{HE} ground truth from that challenge, which was also obtained through a majority vote by the same pathologists.

For the second track, for which a completely new test set was created for this challenge, we used all \ac{MF} annotations of the hotspot \acp{ROI} of the first track. The annotations were done in accordance with the guide for identification of \acp{AMF} by Donovan \etal~\cite{donovan2021mitotic}. For all confirmed \acp{MF}, cell patches (30 microns) were cropped from the image and independently shown to two pathologists (CAB, VW). In case of disagreement, a third expert (TAD) again rendered the final decision.

\begin{figure*}[ht]
    \includegraphics[width=\textwidth]{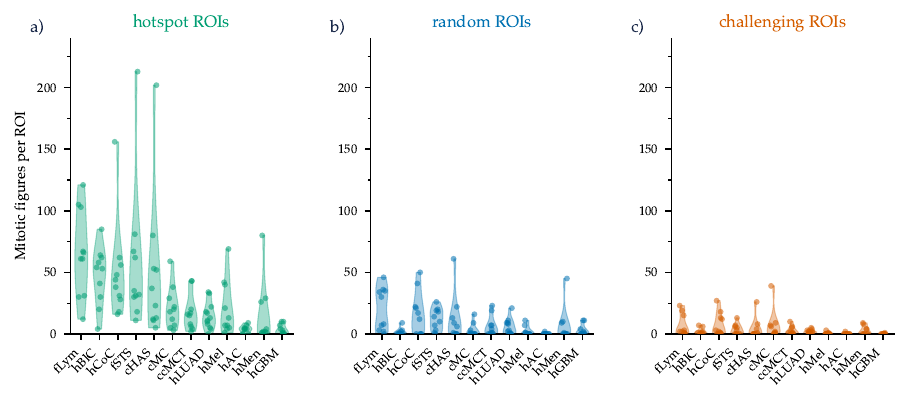}
    \caption{Violin plot showing the distribution of mitotic figures (MFs) in the MIDOG 2025 test set for track 1, with tumor types being sorted by median hotspot MF count. a) The hotspot regions of interest (ROIs) were selected by a pathologist in the most mitotically active tumor region, whereas the random ROI b) regions were sampled from the tissue area of the whole slide image. The challenging regions (c), selected based on the presence of MF imposters, contain the least MF annotations, followed by the random regions and the hotspot ROIs. For the tumor type abbreviations, please consult Table~1.}
    \label{fig:testset-stats-t1}
\end{figure*}

\begin{figure}
\includegraphics[width=0.5\textwidth]{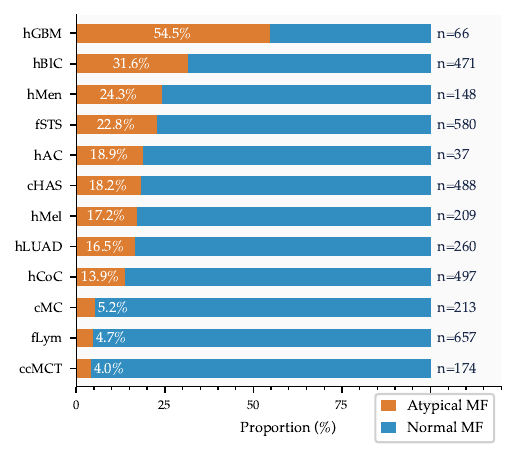}
\caption{Class distribution for all tumor types of the track 2 test set. For the tumor type abbreviations, please consult Table~1. }
\label{fig:distrib-t2}
\end{figure}

\subsection{Dataset statistics of the test set}
The number of \acp{MF} varies considerably across tumor types (Fig.~\ref{fig:testset-stats-t1}), with the feline GI lymphoma (fLym) having the highest overall count (N=657) and the human glioblastoma (hGBM) having the lowest overall count (N=36), reflecting the distinct mitotic activity of the different tumor types.  

For track 2 (\ac{AMF} classification), the atypical label was most often assigned for \acp{MF} in human glioblastoma (hGBM), and least often for canine cutaneous mast cell tumor (ccMCT), as shown in Figure \ref{fig:distrib-t2}. We found Cohen's $\kappa$ of 0.48 between both initial raters for track 1 and of 0.68 for track 2, indicating a moderate and substantial agreement, respectively.

\subsection{Evaluation methods and metrics}
As in the previous challenges, the MIDOG 2025 challenge used the micro-averaged $F_1$ score as primary metric. It is defined as 
\begin{equation}
    F_1 = \frac{2\mathrm{TP}}{2\mathrm{TP}+\mathrm{FN}+\mathrm{FP}}
\end{equation}
where TP, FP, and FN represent the cumulative sum across all \acp{ROI} of true positives, false positives, and false negatives, respectively. The matching between ground truth \acp{MF} and detections was determined using the Hungarian method \cite{kuhn1955hungarian} with a maximum distance of 7.5 micrometers (approximate size of a nucleus). Multiple detections were considered false positives. 
This metric was chosen since it combines recall and precision in a single metric. For the \ac{MC}, an overestimation or underestimation can similarly impact patient prognostication, which is why this combination is sensible. The metric is the most commonly used metric for \ac{MF} performance assessment, allowing comparison to earlier works. We have calculated $F_1$ for the overall dataset, as well as for each tumor type, each \ac{ROI} type, and each combination of tumor type and \ac{ROI} type across all corresponding images.

As an additional secondary metric, we calculated the area under the free-response receiver operating characteristic curve (FROC-AUC). The FROC curve is calculated by varying the detection threshold and determining the recall versus the mean number of false positives per image. To compute the FROC-AUC, sensitivity and mean false positives per image were evaluated across 40 linearly spaced detection thresholds spanning the range of predicted confidence scores. The resulting operating points were interpolated onto 50 uniformly spaced evaluation points in the interval $[0,8]$ false positives per image using linear interpolation, and the area under the interpolated curve was computed via the trapezoidal rule. The upper limit of 8 false positives per image was chosen following~\citet{liu2017detecting}, and reflects a clinically motivated operating range for computer-aided detection systems. A higher FROC-AUC indicates better sensitivity across the full range of operating thresholds at a given false positive budget, and, unlike the $F_1$ score, is independent of a specific detection threshold. As such, it captures a complementary aspect of algorithm performance: where $F_1$ reflects threshold-calibrated precision and recall at a single operating point, FROC-AUC integrates sensitivity over the full operating range, making it sensitive to morphological recognition capability independently of threshold calibration.

Furthermore, as in the predecessor challenge, we calculated \ac{AP} for all participants. \ac{AP} can be understood as an approximation of the area under the precision recall curve, and was calculated as the mean precision for 101 linearly spaced recall values between 0 and 1. We used the implementation provided by the \texttt{torchmetrics} package \cite{Detlefsen2022} (Version 1.9.0). 

For the second track, the main metric was \ac{BA}. As shown in Figure~\ref{fig:distrib-t2}, \ac{AMF} classification is a highly imbalanced problem, with \acp{AMF} being in the minority for almost all tumor domains. The \ac{BA} is defined as average recall across all classes, and spans values between 0.5 (chance) to 1.0 (perfect classification). We additionally calculated the area under the receiver operating characteristic curve (ROC AUC), a metric commonly used in classification performance assessment.  

Moreover, we investigated the rank stability of the challenge, as influenced by the region type (overall, hotspot, random, challenging). We further performed correlation analysis on the metrics across the \ac{ROI} types.

\subsection{Post-Challenge Analysis of Algorithmic Approaches}
A commonality in most challenges is that the submitted approaches differ in a multitude of important algorithmic decisions and components, making it hard to generate insights from a post-challenge analysis. While common success patterns seem to be emerging in pattern recognition challenges, largely based on observation of and comparison of successful and less successful approaches, this is by no means a direct factorial analysis. For instance, the fact that many successful approaches utilize popular techniques for model improvement such as \ac{TTA} or ensembling \cite{eisenmann2023winner} is an indication that these techniques are contributing to model performance, but it could also be that it is a popular pattern that has emerged in the community that is only believed to provide true advantages but possibly only provides marginal benefits, while significantly increasing computational requirements. 

\begin{figure}
    \centering
    \includegraphics[width=\linewidth]{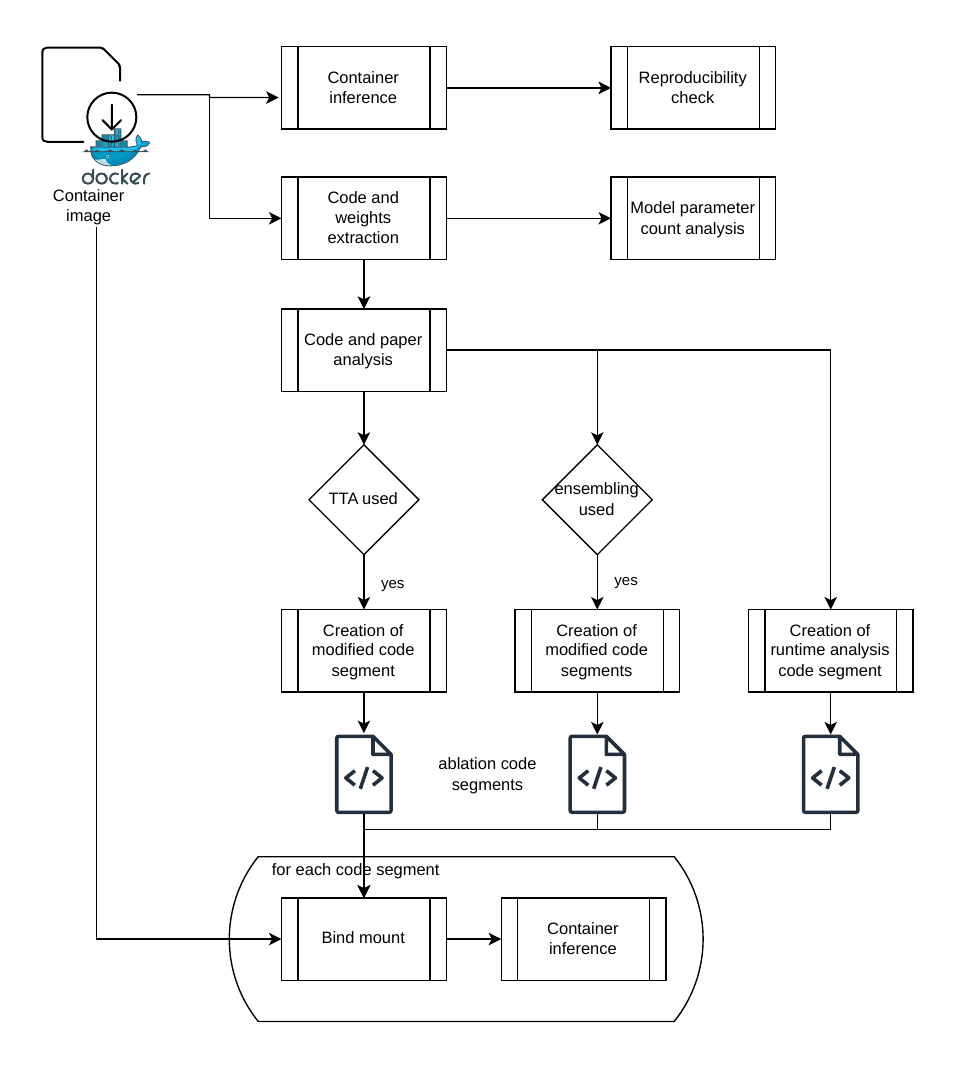}
    \caption{Post-challenge analysis workflow. Model parameter count and inference time were established using modification of the submitted docker containers. Additionally, we ablated the containers from using test time augmentation and ensembling to investigate the effects of both methods.}
    \label{fig:post-challenge-analysis}
\end{figure}
In an attempt to perform a true factorial analysis, we conducted an ablation study on the submitted containers (Figure~\ref{fig:post-challenge-analysis}). For this, we asked the participants for their permission for a post-challenge analysis, and for providing the original container. The vast majority of participants agreed, yielding to post-challenge analysis results on 13/14 submissions for Track~1 and 18/20 for Track~2 (see Tables \ref{tab:t1_full}, \ref{tab:t2_full}). First, the code in each container was manually extracted by overwriting the entrypoint with a shell and providing an external bind mount to the file system of the host, allowing for easy copying between host and container. Subsequently, we performed a manual analysis of the code that was provided by each team, and aligned this against the proceedings paper describing the approach to identify potential discrepancies. In doing so, we identified all model weight files and how they were loaded in the inference pipeline, allowing for a comprehensive assessment of the total parameter count by each approach. 

Finally, we ran ablation studies on the influence of \ac{TTA} and ensembling and performed an inference time analysis. We created modified functional blocks and injected them into the original containers by bind mounting them to replace the original files in the docker container. 

For the inference time analysis, we placed time traces after the loading of the model and before the inference function, and a second one after inference and post-processing was completed. In comparison to the evaluation of the entire call to the docker image, this strips factors such as model loading time and container loading time, which are part of the time assessment in grand-challenge but typically not of interest for algorithmic efficiency comparisons.  We ensured fair comparisons by executing all containers on the same system (a GPU workstation with NVIDIA RTX 4090 GPU, AMD Ryzen 9 7900X 12-Core Processor, and 128GB of RAM) and having no other task run in parallel. For this assessment, we randomly selected 50 data items from the test set (50 ROIs for task 1, and 50 cropout stacks of 16 images for task 2), which we ran on all containers. This also allowed for an analysis of utilized GPU memory. In this analysis, we measured the base GPU memory usage (drawn by utilizing a graphical user interface on the workstation) and ensured that no other relevant processes were active on that machine. We then ran the inference and continuously measured peak additional VRAM utilization.

Next, we identified in each code base the invocation of (1) \ac{TTA} and (2) ensembling of models, if used by the approach, and modified the respective code block to yield code fragments that disable both enhancements individually. For the disabling of ensembling, we ran inference using a single model for each of the provided models and aggregated the results using averaging of the final metrics. 

For teams employing \ac{TTA}, augmentation was disabled by modifying the inference pipeline to process each image in its original orientation only. We subsequently calculated the difference for all metrics that incurred due to both enhancement techniques. Ablations were feasible for 12 of 17 Track~1 teams and 18 of 21 Track~2 teams who provided consent for re-execution of submitted containers; of these, 6 teams in Track~1 and 8 in Track~2 used ensembling, and 4 and 7 respectively used \ac{TTA}.

\subsection{Statistical analysis}

Associations between continuous variables were assessed using Spearman's rank correlation coefficient ($\rho_s$) for analyses involving ranked or non-normally distributed data, and Pearson's $r$ for analyses of raw continuous scores where normality was assumed.  To compare performance across multiple groups, the Kruskal-Wallis test was employed.
Ninety-five percent confidence intervals for Pearson's $r$ were computed via Fisher's $z$-transformation.  All $p$-values are two-sided; a significance threshold of $\alpha = 0.05$ was applied throughout. No correction for multiple comparisons was applied, as all reported analyses were pre-specified based on the challenge evaluation design; correlation analyses involving 12 tumor domains should be interpreted with awareness of limited statistical power. Statistical analyses were performed in Python using \texttt{scipy.stats}~\citep{virtanen2021scipy}.

\section{Overview of submitted methods}

\subsection{Track 1 - MF Detection}
We found a considerable diversity in the approaches submitted for track 1, in particular in the utilized architectures, but also in dataset use (Table~\ref{tab:t1_full}). 

\paragraph{Main Pattern Recognition Approach.} Most teams decided to go for a primary object detection framework to detect mitotic figures (10/14). Amongst those, the most prevalently used framework was the \ac{YOLO} family of models, where various versions were used (YOLOv5, YOLOv8~\cite{varghese2024yolov8}, YOLOv10~\cite{wang2024yolov10}, YOLOv11~\cite{khanam2024yolov11} and YOLOv12~\cite{tianyolov12}) by five teams. Other recent object detection frameworks such as RTMDet~\cite{lyu2022rtmdet}, DETR~\cite{carion2020end}, and even older architectures like DeepLab\-V3+~\cite{chen2018encoder}, were used as object detectors. Four teams decided to frame the task as semantic segmentation task and used flavors of U-Net~\cite{ronneberger2015u} architectures such as nnUNET~\cite{isensee2021nnu}, VM-UNET~\cite{ruan2024vm}, or plain vanilla U-Net~\cite{nasir2025mitodetect}. Most participants used a single stage detector, only four participants used second stage classification networks. 

\paragraph{Datasets.} Most of the participants (13/14) used the MIDOG++ dataset~\cite{aubreville2023comprehensive}, provided by the authors of the challenge. Additionally, 10 out of 14 used the whole slide datasets of canine cutaneous mast cell tumor (MITOS\_\-WSI\-\_CCMCT~\cite{bertram2019large}) and canine mammary carcinoma (MITOS\-\_WSI\-\_CMC~\cite{aubreville2020completely}), which was also provided by the challenge authors. Additionally, contestants used samples from the SPIDER dataset,~\cite{nechaev2025spidercomprehensivemultiorgansupervised}, the NCT-CRC-HE-100K dataset~\cite{kather2019predicting}, the MIDOG 2022 training set~\cite{aubreville2024domain}, a mitosis subtyping dataset by Jahanifar~\cite{jahanifar_2025_15390543}, the OMG Octo dataset~\cite{shen_2024_14246170}, and the alternative label version of the TUPAC16 dataset~\cite{bertram2020pathologist} (see Table~\ref{tab:t1_full}). 

\paragraph{Ensembling and TTA.} Six of the participants used some form of ensembling, either directly in the primary detection/segmentation stage, or in the secondary classification stage. Of note, in the top three, only a single approach used ensembling. Of those approaches that used ensembling, the mean ensemble size was 3.333. \ac{TTA} was used by only four of the participants, including three out of the top five ranked. 

\paragraph{Peer Review Scores.} The participants scored between 9.7 and 14.5 out of 15 in our peer review score. We did not find a significant rank correlation between the peer review score and the rank in the challenge ($rho_s = 0.234, p=0.421$).

\begin{figure*}
    \centering
    \includegraphics[width=\linewidth]{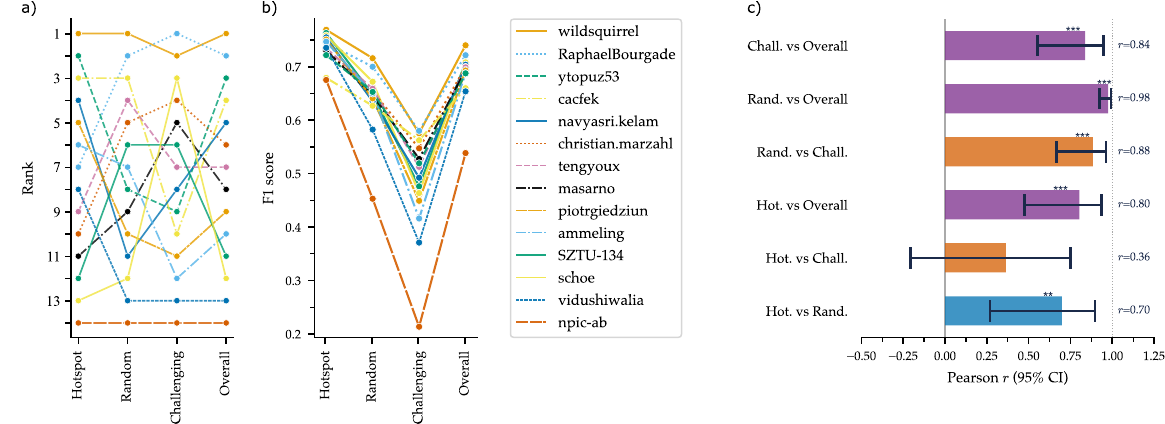}
    \caption{a) Rank and b) $F_1$ scores for all participants across the area types of the track 1 test set. c) shows correlations (Pearson $r$) with 95\% confidence intervals (CI) between the performance of all participants across the various area types (hots.=hotspot, chall.=challenging, rand.=random). The performance in hotspot  and challenging areas had a non-significant correlation, all others were found to be highly significant (** indicates p<0.01, *** indicates (p<0.001).}
    \label{fig:score_stability}
\end{figure*}

\begin{table*}
\centering
\caption{Track~1 (mitotic figure detection) final leaderboard with algorithmic details. $F_1$, FROC-AUC (up to 8\,FP/image) and average precision (AP) are reported on the final test set. OD = object detection; Seg.\ = segmentation. Cla.\ = classification}
\label{tab:t1_full}
\resizebox{\textwidth}{!}{%
\small\setlength{\tabcolsep}{3pt}
\begin{tabular}{cllllrlllrccccl}
\toprule
Rank & Username & Paper & Approach & Architecture & Params & Training data & TTA & Ensemble  & Time (s) & $F_1$ & FROC-AUC & AP & Review \\
\midrule
1 & wildsquirrel  & Lv et al.~\cite{lv2026kongnetmultiheadeddeeplearning} & Seg. (disc) &  KongNet~\cite{lv2026kongnetmultiheadeddeeplearning} & 381M & M++,CMC,CCMCT,\cite{jahanifar_2025_15390543} & Yes & 3 models & 15.0 & \textbf{0.740} & \textbf{4.195} & \textbf{0.608} & 10.0 \\
2 & RaphaelBourgade & Bourgade et al. ~\cite{proc_Bourgade2026}& OD & YOLOv12 (yolo12m)~\cite{tianyolov12} & 20M & M++,CMC,CCMCT & No & No & 14.9 &\textbf{0.722} & \textbf{4.932} & \textbf{0.634} & 13.0 \\
3 & ytopuz53 & Topuz et al.~\cite{proc_Topuz2026} &  OD & SDF-YOLO & 18M & M++,CMC,CCMCT & Yes & No & 4.0  & \textbf{0.708} & \textbf{5.347} & \textbf{0.733} & 13.5  \\
4 & cacfek & Fekete et al.~\cite{shen2025pancancermitoticfiguresdetection} & OD & Yolov10x~\cite{wang2024yolov10} & 159M &M++,CMC,CCMCT,~\cite{shen_2024_14246170},~\cite{aubreville2024domain} & Yes & 5 models & 20.0 & 0.706 & {5.150} & {0.660} & 11.0  \\
5 & navyasri.kelam & Kelam et al.~\cite{proc_Kelam2026} & OD & Yolov8~\cite{varghese2024yolov8} + yolov5mu & 79M & M++,CMC,CCMCT & No & 2 models & 3.1 & 0.704 & 5.221 & 0.673 & 10.0  \\
6 & christian.marzahl & Marzahl et al.~\cite{proc_Marzahl2026} & OD & RTMDet\_s\_1912\_1192~\cite{lyu2022rtmdet},  & 36M & M++,CMC,CCMCT,\cite{bertram2020pathologist},~\cite{nechaev2025spidercomprehensivemultiorgansupervised},~\cite{kather2019predicting} & No & 4 models &  11.7 &  0.700 & 5.303 & 0.740 & 14.5 \\
7 & tengyoux & Xu et al.~\cite{proc_Xu2026} & Seg. + Cla. & nnUnetV2~\cite{isensee2021nnu}, EfficientNet-b3/-b3/v2-s~\cite{tan2019efficientnet} & 577M & M++,CMC,CCMCT & Yes & 3 models (Cla) & 35.3 & 0.697 & 4.998 & 0.670 & 11.0  \\
8 & masarno & Percannella et al.~\cite{proc_Percannella2026b} & Seg. & VM-UNET~\cite{ruan2024vm} & 108M & M++ & No & 3 models & 19.3 &  0.694 & 4.771 & 0.587 & 13.0  \\
9 & piotrgiedziun & Giedziun et al.~\cite{proc_Giedziun2026b} & OD & RF-DETR~\cite{sapkota2025rfdetrobjectdetectionvs} & 32M & M++,CMC,CCMCT,\cite{nechaev2025spidercomprehensivemultiorgansupervised} & No & No & 3.8 & 0.693 & 5.187 & 0.656 & 10.0  \\
10 & SZTU-134 & Xiao et al.~\cite{proc_Xiao2026} & OD + Cla. & YOLO 11x~\cite{khanam2024yolov11}, ConvNeXt-Tiny~\cite{liu2022convnet} & 82M & M++,CMC,CCMCT & No & No & 7.2 & 0.688 & 4.999 & 0.623 & 13.5  \\
11 & schoe & Choe et al.~\cite{proc_Choe2026} & Seg. & U-Net~\cite{ronneberger2015u} & 49M & M++,CMC,CCMCT,\cite{gamper2019pannuke},\cite{bertram2020pathologist} & No & No & 11.6 & 0.660 & 4.105 & 0.507 & 13.5  \\
12 & vidushiwalia & Walia et al.~\cite{proc_Walia2026} & OD & DETR~\cite{carion2020end} & 42M & M++,CMC,CCMCT & No & No &  3.5 & 0.654 & 4.784 & 0.681 & 11.5  \\
13 & krhasan02 & Hasan et al.~\cite{rakib_hasan_2025_17017921} & OD & DeepLab V3+~\cite{chen2018encoder} & -- $^1$ & M++ & No & No &  -- $^1$ & 0.634 & 4.243 & 0.487 & 9.7  \\
14 & npic-ab & Broad et al.~\cite{proc_Broad2026} & OD + Cla. & FCOS~\cite{tian2019fcos}, VGG19~\cite{simonyan2014very} & 222M & M++ & No & No & 5.4 & 0.538 & 3.967 & 0.587 & 9.7 \\
%15 & gdeotale123 & N/A $^2$& -- $^2$ & -- $^2$ & -- $^1$ & -- $^2$ & --$^2$  & --$^2$ & 0.368 & 2.280 & 0.364 & -- \\ DISQUALIFIED - No preprint
% 16 & procrastinator31 & Song et al.~\cite{song2025challengeslessonsmidog2025} & OD + Cla. & FasterRCNN + 3 CNN Ensemble & --$^1$ & M++, CMC, CCMCT & Yes & Yes & 0.468 & 0.426 & 0.103 & -- \\ DISQUALIFIED - No preprint
%17 & rishabanand & N/A $^2$ & -- $^2$ & -- $^2$ & -- $^2$ & -- $^2$ & Yes & Yes & 0.063 & 0.308 & 0.024 & -- &\\.  DISQUALIFIED - No preprint
\midrule
baseline & ammeling  & Banerjee et al.~\cite{proc_Banerjee2026} & OD & FCOS~\cite{tian2019fcos} & 95M & M++ & No & No & 6.4 & 0.688 & 5.139 & 0.732 & -- \\
\bottomrule
\end{tabular}}
\footnotesize Training data: M++=MIDOG++~\cite{aubreville2023comprehensive}, CMC=MITOS\_WSI\_CMC~\cite{aubreville2020completely}, CCMCT=MITOS\_WSI\_CCMCT~\cite{bertram2019large},  Params: Aggregated model parameter count. Time: mean inference time per ROI on RTX~4090. Review: mean peer review score (max.~15). Top-3 results in bold. $^1$~Model container was not provided. 
\end{table*}

\subsection{Track 2 - AMF Classification}

The key differences between submissions for Track 2 were related to model training and the ensembling strategy. (see Table \ref{tab:t2_full}):
\paragraph{Model Training.} 
Almost no participants chose to train a model from scratch: 7 out of 20 participants chose to use \ac{LoRA} as parameter-efficient adaptation of foundation models and 10 out of 20 participants chose to use fine-tuning of other pre-existing models. Two teams used distillation as primary learning paradigm.
As feature extractors, most participants used architectures employing convolutional layers, such as EfficientNet~\cite{tan2019efficientnet}, ConvNeXt~\cite{liu2022convnet}, or ResNet~\cite{he2016deep}.  For \ac{LoRA}, seven participants chose recent pathology foundation models, such as UNI~\cite{chen2024towards}, Virchow~\cite{vorontsov2023virchow,zimmermann2024virchow2} or HIBOU~\cite{nechaev2024hibou}. The winning team (Balezo et al.~\cite{proc_Balezo2026}) chose to low-rank adapt the general-purpose foundation model DINOv3~\cite{simeoni2025dinov3}. One submission (Qi et al.~\cite{proc_Qi2026}) used fine-tuning of a general-purpose vision transformer model pre-trained on ImageNet (EfficientVit~\cite{liu2023efficientvit}). 

\paragraph{Ensembling.} Another frequently employed strategy was ensembling of heterogeneous models: 12 out of 20 teams used multiple models and ensembled their outputs for the prediction. The mean size of the ensemble was 4.667. In the most notable case of ensembling, Ochi et al.~\cite{proc_Ochi2026}, used a combination of three foundation models (Virchow~\cite{vorontsov2023virchow}, Virchow2~\cite{zimmermann2024virchow2}, UNI~\cite{chen2024towards} and an ImageNet-finetuned ConvNeXt V2~\cite{woo2023convnext}, fused by the autogluon framework~\cite{erickson2020autogluon} using a combination of 438 models. Of note, the winning approach was the only in the top five that did not use any model ensembling. 

\paragraph{Datasets.} All approaches used the challenge's training set~\cite{weiss_2025_16044804}. The majority (12/20) also used the smaller breast cancer atypical mitosis dataset AMi-Br~\cite{bertram2025histologic}. During the challenge, the challenge participants Shen et al. made available an additional dataset, OMG-Octo Atypical, containing 1,378 atypical mitotic figures~\cite{shen2025omg}. This dataset was reported to have been used by 8 out of 20 participants. Besides this, additional datasets used for training include an atypical subtyping datset by Jahanifar et al.~\cite{jahanifar_2025_15390543}, stMIDOG/LUNG-MITO~\cite{ivan2026subphase}, and GBM-TCGA\cite{liu2025deep}.

\paragraph{Peer Review Scores.} The participants reached an averaged peer review score across reviewers of 8.0 to 15.0. We found no significant rank correlation between rank in the challenge and peer review scores ($\rho_S =-0.270,p=0.250$).

\begin{table*}[ht]
\centering
\caption{Track~2 (atypical mitotic figure classification) final leaderboard with algorithmic details. Balanced accuracy and ROC AUC are reported on the final test set.}
\label{tab:t2_full}
\small\setlength{\tabcolsep}{3pt}
\resizebox{\textwidth}{!}{%
\begin{tabular}{cllllrlllrccll}
\toprule
Rank & Username & Paper & Adaptation & Architecture & Params & Training data & TTA & Ensemble & Time (s) & Bal.\ Acc. & ROC AUC & Review  \\
\midrule
1 & guillaume.balezo & Balezo et al.~\cite{proc_Balezo2026} & LoRA & DINOv3-H~\cite{simeoni2025dinov3} & 842M & M25,AMi-Br,OMG & Yes & No & 0.421 & \textbf{0.908} & \textbf{0.970} & 13.0  \\
2 & nasires & Nasir et al.~\cite{proc_Nasir2026} & LoRA & Virchow2~\cite{zimmermann2024virchow2} & 631M & M25,AMi-Br,OMG,~\cite{jahanifar_2025_15390543} & Yes & 3 models & 11.450 & \textbf{0.901} & \textbf{0.967} & 10.0  \\
3 & yohsuke.yamagishi & Yamagishi et al.~\cite{proc_Yamagishi2026} & FT & ConvNext-v2-base~\cite{woo2023convnext} & 438M & M25 & No & 5 models & 1.247 & \textbf{0.900} & \textbf{0.971} & 13.0  \\
4 & qixuan1234 & Qi et al.~\cite{proc_Qi2026} & FT & EfficientViT-L2\cite{liu2023efficientvit} & 320M$^{1,2}$ & M25,Ami-Br & Yes & 5 models & --$^1$ & 0.897 & 0.962 & 9.5  \\
5 & masarno & Percannella et al.~\cite{proc_Percannella2026} & FT & Custom Hover-Net\cite{graham2019hover} & 447M & M25 & Yes & 5 models & 0.686 & 0.897 & 0.962 & 12.5  \\
6 & zerostarcraft & Meng et al.~\cite{proc_Meng2026} & VPT & UNI2-h~\cite{chen2024towards} & 682M & M25 & Yes & No & 2.879 & 0.896 & 0.962 & 14.5  \\
7 & krausara & Krauss et al.~\cite{proc_Krauss2026} & FT & ConvNeXtBase~\cite{liu2022convnet} & 793M & M25,AMi-Br,OMG & No & 3 models & 22.534 & 0.889 & 0.960 & 15.0  \\
8 & lrc9859 & Ramchandani et al.~\cite{proc_Ramchandani2026} & LoRA & Virchow-Base~\cite{vorontsov2023virchow} & 647M & M25,AMi-Br,OMG & No & 3 models & 0.986 & 0.884 & 0.958 & 11.5  \\
9 & be\_yuan & Ochi et al.~\cite{proc_Ochi2026} & LoRa/FT & UNI, Virchow, Virchow2, ConvNext-V2 & 1658M$^3$ & M25,AMi-Br,OMG,\cite{jahanifar_2025_15390543} & No & 4 models & 0.815 & 0.884 & 0.946 & 11.5  \\
10 & piotrgiedziun & Giedziun et al.~\cite{proc_Giedziun2026} & LoRA & Virchow2-Base~\cite{zimmermann2024virchow2} & 708M & M25,AMi-Br,\cite{ivan2026subphase} & Yes & 13 models & 4.859 & 0.882 & 0.202 & 12.0  \\
11 & cacfek & Shen et al.~\cite{shen2025pancancermitoticfiguresdetection} & FT & ConvNext~\cite{liu2022convnet} & 139M & M25,AMi-Br,OMG,\cite{liu2025deep},\cite{ivan2026subphase},\cite{liu2025deep} & No & 5 models & 0.232 & 0.878 & 0.953 & 11.0  \\
12 & schoe & Choe et al.~\cite{proc_Choe2026} & Distillation & UNet\cite{ronneberger2015u} & 173M & M25,AMi-Br,OMG & No & No & 0.189 & 0.872 & 0.948 & 13.5  \\
13 & saipradeepvg & Kotte et al.~\cite{proc_Kotte2026} & FT & ResNet50~\cite{he2016deep} & 25M & M25 & No & No & 0.202 & 0.869 & 0.944 & 10.5  \\
14 & navyasri.kelam & Kelam et al.~\cite{preprint_kelam_2025} & LoRA & UNI~\cite{chen2024towards} & 307M & M25,AMi-Br & No & No & 0.236 & 0.868 & 0.944 & 8.0  \\
15 & Leire & Benito-Del-Valle et al.~\cite{proc_BenitoDelValle2026} & FT & ConvNext-small~\cite{liu2022convnet}& 247M & M25 & Yes & 5 models & 0.299 & 0.868 & 0.964 & 11.0  \\
16 & Kaustubh\_Atey & Atey et al.~\cite{proc_Atey2026} & Distillation & DenseNet-121~\cite{huang2017densely} & 7M & M25,AMi-Br,OMG & No & No & 0.294 & 0.859 & 0.957 & 12.0  \\
17 & mirazzak & Dukre et al.~\cite{proc_Dukre2026}& FT & DenseNet-121~\cite{huang2017densely} & 7M & M25 & No & No & 0.258 & 0.850 & 0.927 & 11.0  \\
18 & mlafarge & Lafarge et al.~\cite{proc_Lafarge2026} & -- $^4$ & Custom ResNet-41~\cite{lafarge2020roto} with P4M GroupConv~\cite{cohen2016groupCNN} & 2M & M25,AMi-Br & No & No & 0.471 & 0.829 & 0.905 & 11.0  \\
19 & chillice & Xu et al.~\cite{proc_Xu2026} & FT & EfficientNet-b3/-b5~\cite{tan2019efficientnet}, and InceptionV3~\cite{szegedy2015going} & 61M & -- & Yes & 2 models & 2.330 & 0.824 & 0.955 & 11.0  \\
20 & sercan.cayir & \c{C}ayir et al.~\cite{proc_Cayr2026} & LDA / CatBoost & HIBOU-L~\cite{nechaev2024hibou} and Barlow-Twins~\cite{zbontar2021barlow} & -- $^1$ & M25 & No & 3 models & -- $^1$ & 0.671 & 0.759 & 12.0  \\
\midrule
Baseline  & maubreville & Banerjee et al.~\cite{proc_Banerjee2026} & FT & EfficientNet-2M~\cite{tan2021efficientnetv2} & 53M & M25 & No & No & 0.265 & 0.827 & 0.907 & --  \\
\bottomrule
\end{tabular}}
\footnotesize Approach: FT=fine-tuning of ImageNet-pretrained model, LoRA: Low-rank adaptation, VPT: visual prompt tuning. Training data: M25=MIDOG~2025 atypical set~\cite{weiss_2025_16044804},  AMi-Br=AMi-Br atypical dataset~\cite{bertram2025histologic}, OMG=OMG-Octo atypical dataset~\cite{shen2025omg}. Params: single model parameter count. Time: mean inference time per ROI on RTX~4090.  Review: mean peer review score (max.~15). Top-3 results in bold. $^1$ Participant did not provide docker container. $^2$ Value taken from literature. $^3$ Model fusion not counted. $^4$ No fine-tuning or adaptation was mentioned in the proceedings paper.
\end{table*}

\begin{figure*}
    \includegraphics[width=\textwidth]{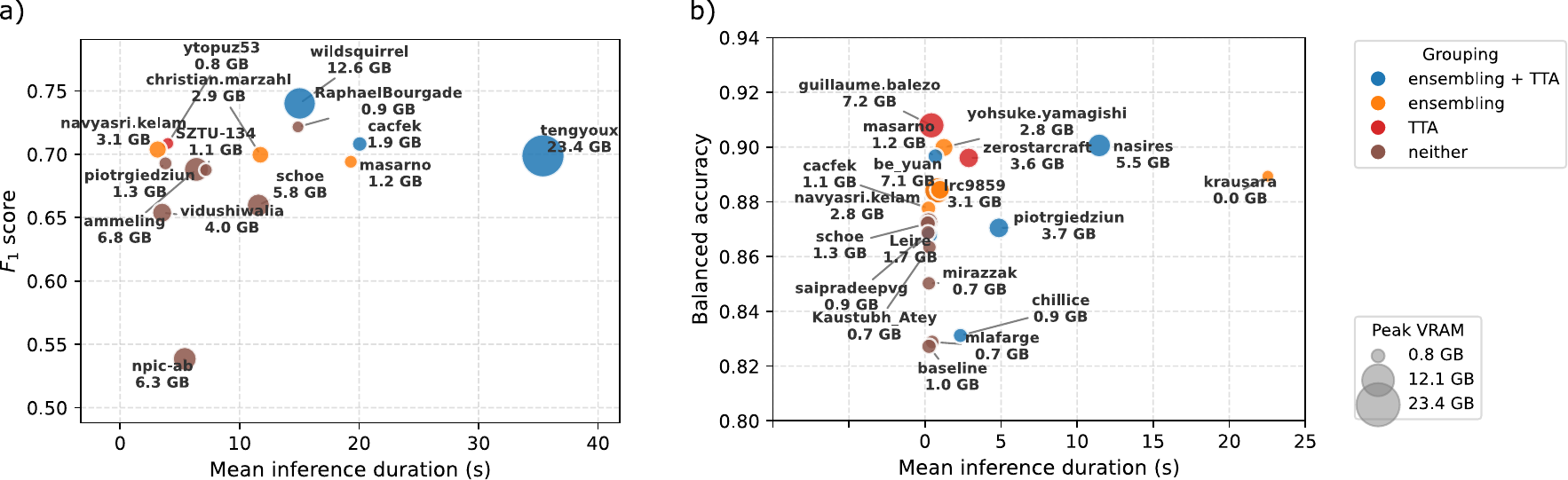}
    \caption{Inference time vs. main metric for track 1 (a) and track 2 (b) of the MIDOG 2025 challenge. Inference time was determined on a Linux workstation with an NVIDIA RTX 4090. Bubble size indicates max. VRAM usage of container, averaged over 50 cases.}
    \label{fig:inference_time}
\end{figure*}

\section{Results}
The main results are included in Tables \ref{tab:t1_full} and \ref{tab:t2_full}, and will be discussed in the following.

\subsection{Track 1 - Detection}
The highest result in the main challenge metric in Track~1 ($F_1$ score) was achieved by Lv et al.~\cite{lv2026kongnetmultiheadeddeeplearning,nasir2025mitodetect}, reaching an overall $F_1$ score of 0.740 (see Table \ref{tab:t1_full}). The runner-up for this track were Bourgade et al.~\cite{proc_Bourgade2026}, reaching an overall $F_1$ score of 0.722. In the secondary challenge metric, FROC AUC, the leading team were Topuz et al.~\cite{proc_Topuz2026}, reaching a score of 5.347. The runner-up for FROC AUC were Marzahl et al.~\cite{proc_Marzahl2026} with a score of 5.303. 

\paragraph{Score Stability.} We evaluated the rank stability across the different \ac{ROI} types (hotspot, random, challenging, and overall) in rank across the field of the challenge participants. As shown in Figure \ref{fig:score_stability}a), the rank strongly depended on the \ac{ROI} type. As Figure \ref{fig:score_stability}b) shows, the variance in $F_1$ score is small in the hotspot regions. In consequence, even small changes in $F_1$ in hotspot regions can lead to a change in rank. We furthermore found significant correlations between the overall $F_1$ score and all individual \ac{ROI} types (see Fig. \ref{fig:score_stability}c). However, with r=0.36 (CI95=[-0.206, +0.750], p = 0.201) there was no significant correlation between hotspot $F_1$ score and the same metric in challenging areas, highlighting that good hotspot predictions models are not necessarily good predictors for challenging areas.

\paragraph{Tumor Type Dependency.} We found a considerable dependency on the tumor type, with $F_1$ scores ranging from 0.444 (human glioblastoma) to 0.808 (canine hemangiosarcoma) in median for all participants (see supplementary Figure \ref{fig:f1_score_by_tumor_type}). For the challenging areas, this difference was even more pronounced, with the minimum achieved likewise on human glioblastoma (0.010) and the maximum achieved on feline lymphoma (0.634). Using a Kruskal-Wallis-Test, we evaluated the significance of tumor type, pooled across all area types, and found it to be significant (p < 0.0001). We did not conduct individual pairwise tests.

\paragraph{ROI Type Dependency.} We investigated how strongly results varied across the different area types of our challenge (overall, hotspot, random, challenging). In the main challenge metric for track 1 ($F_1$ score), we found a mean overall $F_1$ score of 0.685. In hotspot areas, this was considerably elevated, yielding a mean $F_1$ of 0.735 across participants (see also Fig. \ref{fig:score_stability}b). In random areas, the performance dropped to a mean value across participants of $0.638$. In challenging areas, this effect was even more pronounced, yielding a mean $F_1$ score of only 0.479 (difference to hotspot areas: 0.256). 

The drop in $F_1$ score can largely be attributed to a drop in precision (see supplementary Figure \ref{fig:precision-recall-areatypes}). Compared to the precision value of all \acp{ROI} combined (0.704), hotspot \acp{ROI} had the highest score of 0.805 and performance markedly decreased in random (0.614) and challenging areas (0.400), indicating an increased incidence of false positives in the latter. The difference in precision between hotspot and challenging areas was 0.405, translating into an increased false detection rate by approximately 208\%. The detrimental effect of a different area type was less pronounced in the recall value: We found an overall mean recall value of 0.681 across participants, and recall values of 0.682, 0.685, and 0.662 for the hotspot, random, and challenging areas, respectively. For a list of per-tumor-type and per-team precision and recall values, as well as the $F_1$ and FROC AUC values, please consult supplementary Figures \ref{fig:precision-recall-participants} and \ref{fig:f1-froc_auc-participants}.

\subsection{Track 2 - Classification}
The highest performance in the main challenge metric (\ac{BA}) was achieved by Balezo et al.~\cite{proc_Balezo2026} with a value of 0.908. In the secondary ROC AUC metric, Yamagishi et al.~\cite{proc_Yamagishi2026} achieved a marginally better result (0.971 vs. 0.970 of Balezo et al.). The runner-up in \ac{BA} were Nasir et al.~\cite{proc_Nasir2026} with a balanced accuracy of 0.901, only narrowly beating the third-ranked Yamagishi et al. \cite{proc_Yamagishi2026}, who achieved a score of 0.900. Overall, the top three teams demonstrated remarkably strong performance, with a difference of less than one percentage point in \ac{BA} separating the first from the third place. 

\paragraph{Tumor Type Dependency.}
As shown in the supplementary Figure \ref{fig:ba_roc_auc_tumor_type_t2}, there was a considerable variance across the domains (tumor types) of the test set. The difference was assessed to be significant using a Kruskal-Wallis test for both \ac{BA} (H=98.757, p=0.0000) as well as ROC AUC (H=78.752, p=0.0000). The lowest median \ac{BA} was achieved for feline lymphoma (fLym) and the highest median \ac{BA} was achieved for human astrocytoma (hAC). As shown in supplementary Figure \ref{fig:t2_ss_by_domain}, the classification of \acp{AMF} in feline lymphoma was mostly restricted by a low recall.

\subsection{Reproducibility analysis}
%While the participants had submitted dockerized algorithms to the grand-challenge platform during the challenge, these were not accessible to the organizers in retrospect. Therefore, for an independent post-challenge analysis, we asked the participants to provide the organizers with their respective container image. The vast majority of participants complied with this request. 
We re-evaluated all containers provided by the participants to the organizers on the test set and found the containers to reproduce the grand-challenge official challenge results with sufficient precision ($F_1$: deviation mean absolute $3.56 \cdot 10^{-5}$, max absolute: 0.18 percentage points, balanced accuracy: mean absolute $2.24 \cdot10^{-5}$, max absolute 1.13 percentage points).

\subsection{Inference time and memory analysis}
Participants had a five minute time limit on grand-challenge for each inference job. However, inference jobs include starting of the docker container, loading of the model, model inference and post-processing, and can be subject to secondary load on the system running the inference task. The actual time available for inference was thus much shorter, and, in track 1, was furthermore subject to image size and post-processing time additionally scaled with detection results due to non-maximum suppression having typically an $\mathcal O (n^2)$ complexity. \ac{VRAM} was limited to 16 GB in the grand-challenge environment and to 24 GB in the post-challenge evaluation. Inference time varied strongly across the field of participants (see Fig. \ref{fig:inference_time}).

In Track 1, the approach using the highest average inference time per \ac{ROI} (35.386s) was by Xu et al.~\cite{proc_Xu2026}, which used a two-stage approach based upon nnUnet \cite{isensee2021nnu}, a three model second stage for classification and a final ensembling mechanism by a random forest, combined with \ac{TTA}. The team with the lowest inference time for track were Kelam et al.~\cite{proc_Kelam2026}, who only used 3.515s of inference time on average and still achieved the fifth place on the leader board. We found no significant rank correlation between inference time and $F_1$ score ($\rho_S =0.367$, $p=0.196$) or peak \ac{VRAM} usage and $F_1$ score ($\rho_s=-0.253$, $p=0.383$). 

For track 2, we found a similarly high variance of inference time and \ac{VRAM} usage, as shown in Figure \ref{fig:inference_time}b. On average, the GPU \ac{VRAM} use was much lower in track 2 (2.424\,GB) compared to track 1 (5.151\,GB). With 22.534s, Krauss et al.~\cite{proc_Krauss2026} had the longest inference time per batch of 16 images, however, without using the GPU (Fig. \ref{fig:inference_time}). The shortest time was achieved by Choe et al.~\cite{proc_Choe2026} with 0.189s. For track 2, we found a significant rank correlation between peak VRAM usage and balanced accuracy ($\rho_s = 0.616$, $p=p=0.005$) but not between inference time and balanced accuracy ($\rho_s = 0.411$, $p=0.080$). 

\subsection{Ablation studies}
\begin{figure}
    \includegraphics[width=\linewidth]{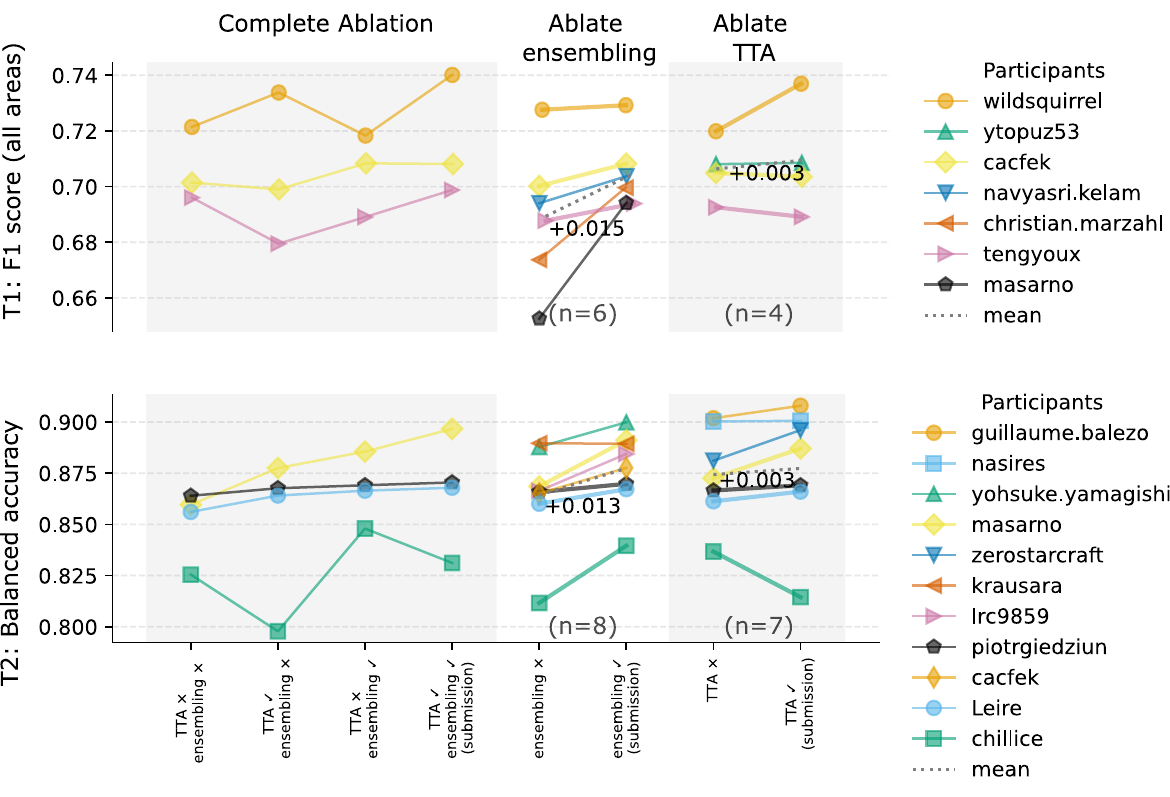}
    \caption{Ablation study for test-time augmentation (TTA) and ensembling. Shown are only participants that utilized either TTA or ensembling in track one (top) and track two (bottom). }
    \label{fig:ablation}
\end{figure}

Our component analysis of \ac{TTA} or ensembling reveals a mixed benefit of both methods for both tracks (Figure~\ref{fig:ablation}). By using ensembling, we found the participants of Track 1 had a mean increase of the $F_1$ metric on the entire test dataset of 1.549 percentage points, and a median increase of 0.886 percentage points. In the hotspot \acp{ROI} alone, we found an increase of 1.404 percentage points in mean and 1.290 percentage points in median. In the random areas, we found a mean increase of 1.661 percentage points (1.198 in median). In the challenging areas, the effect was even more pronounced, leading to an increase of, in mean and median, 2.212 and 1.691 percentage points, respectively. We can thus summarize that the effect of ensembling grew with the data distribution moving further outside of the training distribution. \ac{TTA} had a less pronounced effect on the main challenge metric, with only 0.320 percentage points improvement in mean and almost no change in median (-0.042 percentage points), with only marginal changes over the \ac{ROI} types.

In the \ac{AMF} classification track, the teams benefited from both techniques in a similar way, yielding a median and mean increase of 1.230 and 1.299 percentage points by ensembling. By \ac{TTA}, the results in the second track increased on average by 0.299 percentage points and 0.472 percentage points in median. 

For four models in Track~2, ablation was not possible: For two, the authors did not provide docker containers for post-challenge analysis (see Table \ref{tab:t2_full}). In one instance (Ochi et al.~\cite{proc_Ochi2026}), a highly complex fusion model of very heterogeneous other models was used, making an ensembling ablation of this model questionable. In another instance (Nasir et al.~\cite{nasir2025mitodetect}), the model ensemble was fused in one singular model, prohibiting ablation.

\section{Discussion}

The \ac{MIDOG} 2025 challenge was the first challenge to ever incorporate testing outside of hotspot regions, and -- with twelve independent tumor domains -- provides the largest and most comprehensive test set to date. 

Even though open training data with annotations in entire WSIs exists and was used by almost all participants, we still see a dramatic loss of performance for random and challenging areas. In the hotspot regions, however, we found overall high performance with only minor variance across participants (see Fig. \ref{fig:score_stability}). Our results hint at a loss of generalization due to a missing data variance in existing datasets. This calls for the curation of datasets with a higher domain variance, particularly with respect to region selection. 
For the deployment of current algorithms, this means that algorithms should not be deployed for WSI usage, unless specifically validated on a wider area selection or on entire \acp{WSI}.

% YOLO was very prevalent in T1, especially in top ranks, foundation models with LoRA adaptation were widely used in T2
In both tracks, the challenge has revealed architectural trends in the highest ranking teams: While the YOLO model family was often chosen for the object detection track, \ac{LoRA}-adaptation of foundation models was a clear trend in the top ranks of the atypical classification track.  

Many of the successful teams employed common machine learning tricks such as ensembling and \ac{TTA}. Ensembling provided consistent benefits across both tracks, however, it is worth noting that the top ranked participant of track 2 (Balezo et al.) did not make use of it. Furthermore, out of the top three methods in track 1, only one used ensembling, suggesting that while ensembling can improve the performance, it is not a prerequisite for achieving top performance. 
In contrast, \ac{TTA} had a much smaller impact on performance and was in some cases even detrimental. Still, \ac{TTA} was used by three out of the top five teams for track 1 and four out of the top five in track 2. While \ac{TTA} is thus a well-known technique and easy to implement, it does add significant computational overhead, and was not effective for this challenge. While broader investigations beyond this challenge are necessary to confirm our observations, this result might serve as an indicator to discourage the use of \ac{TTA} in pathology challenges with current models.

We found no significant correlation between paper quality, as assessed through peer review, and challenge performance for either track. This points towards an important dilemma in challenge workshops: While paper scoring, which is based on innovative methods and an intriguing presentation, is typically used to identify contributions for talks, this strategy might omit submissions with highest model performance. 

A further consideration concerns not only the availability of training data, but the spatial context it provides. Two of the additional datasets used by participants: OMG-Octo~~\cite{shen_2024_14246170} and the Jahanifar mitosis subtyping dataset~\cite{jahanifar_2025_15390543}, provide only 64×64 px patches centered on individual mitotic figures, a constraint partly imposed by clinical data-sharing requirements. While such patches are well suited for the classification task of Track 2, they offer limited spatial context for object detection models (\textit{e.g.} YOLO), where surrounding tissue architecture is a key cue for distinguishing mitotic figures. This may partly explain why teams (see Table \ref{tab:t1_full}) which utilized training data containing broader spatial context performed comparatively well in challenging regions. More broadly, our results suggest that future dataset curation efforts should consider not only diversity in tumor type and scanner, but also the spatial extent of the provided annotations, as this directly affects which architectural classes can benefit from a given resource.

% LIMITATIONS:

We also want to highlight limitations of our work. The annotation process in the two new tumor types (hotspot areas) as well as for all tumor types in the random and challenging areas (i.e., all \acp{ROI} that have not been previously utilized in MIDOG 2022) deviated slightly from the previous MIDOG 2022 workflow. While both annotations methods were supported by \ac{PHH3} IHC to identify \acp{MF} in the first annotation step, there was a difference in the label classes, which could lead to label drift. However, we estimate the impact of this deviation to be low, given that both annotation workflows conducted final decision through a majority vote by three pathologists.

% AP vs FROC-AUC
In the first track of our challenge, we used \ac{AP} as tertiary metric for the threshold-independent algorithms assessment. Although \ac{AP} is well established in object detection, it is sensitive to the initial confidence threshold applied before non-maximum suppression (NMS). Since NMS has a computational complexity of $\mathcal O(n^2)$ and execution time was limited, participants were incentivized to use higher cutoff values to reduce processing time. Conversely, lower cutoffs generate a broader range of precision–recall pairs, extending the precision–recall curve and artificially increasing\ac{AP} as previously observed in the MIDOG 2022 challenge~\cite{aubreville2024domain}. However, this increase does not necessarily reflect improved class discrimination, representing a major limitation of this metric. In contrast, FROC-AUC is constrained by a predefined false-positive rate per image and is therefore less susceptible to inflation from low confidence thresholds.

The MIDOG 2025 challenge was the most successful challenge in terms of participation, attracting more submissions than previous iterations. Our results indicate that, in particular, mitotic figure detection performance in hotspot regions is a task that is carried out with high robustness across unseen domains, which is a precondition for clinical use. The availability of new algorithms for atypical classification, created in the context of this challenge, allows for the deeper investigation of the pathological role of \acp{AMF}.

In summary, the MIDOG 2025 challenge has substantially advanced the field of computational pathology by establishing a new benchmark for mitotic figure detection and atypical mitosis classification across an unprecedented breadth of tumor domains and tissue regions. The challenge highlights both the maturity of current approaches -- particularly for hotspot-based detection -- and the clear gaps that remain when algorithms are applied to the broader, more heterogeneous landscape of whole slide images. The architectural trends identified here, including the dominance of YOLO-based detectors and LoRA-adapted foundation models, provide a valuable snapshot of the current state of the art and a foundation for future methodological development. Moving forward, we advocate for the curation of more diverse training datasets, rigorous validation beyond hotspot regions, and the establishment of evaluation frameworks that reward generalization as much as peak performance. We hope that the datasets and baselines released alongside this challenge will serve as a lasting resource for the community, and that the findings presented here will inform the responsible translation of mitosis detection algorithms into clinical practice.

\subsection*{Acknowledgements}
M.A. and S.B. acknowledge funding by the Deutsche Forschungsgemeinschaft (DFG, project number: 520330054), C.A.B. and V.W. acknowledge funding by the Austrian Research Fund (FWF, project number: I 6555). J.A. acknowledges support by the Bavarian State Ministry of Science and the Arts (project Fokus-TML). K.B. acknowledges funding by the DFG, project number 460333672 CRC1540 EBM.

The MIDOG challenge received financial support from MIRA vision microscopy GmbH, Göppingen, Germany and Single-Cell Technologies Ltd, Szeged, Hungary to cover the platform costs. We furthermore acknowledge support by the MICCAI special interest group (SIG) on computational pathology, who donated monetary prizes for the winners of the challenge.

\subsection*{Organization Team}
The MIDOG 2025 challenge was organized by (in alphabetic order): Jonas Ammeling, Marc Aubreville, Sweta Banerjee, Christof A. Bertram, Katharina Breininger, Dominik Hirling, Peter Horvath, Nikolas Stathonikos, and Mitko Veta.

\appendix
\section{Supplementary Figures}
\renewcommand{\figurename}{Supplementary Figure}
\renewcommand{\thefigure}{S\arabic{figure}}

\setcounter{figure}{0}
\begin{figure}[h!]
    \includegraphics[width=\linewidth]{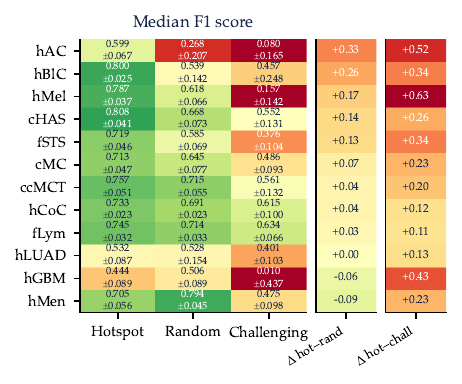}
    \caption{$F_1$ score by tumor type for Track 1 of the challenge (mitotic figure detection). Shown are median values $\pm$ inter-quartile range (25th-75th percentile).}
    \label{fig:f1_score_by_tumor_type}
\end{figure}

\begin{figure*}
    \includegraphics[width=\linewidth]{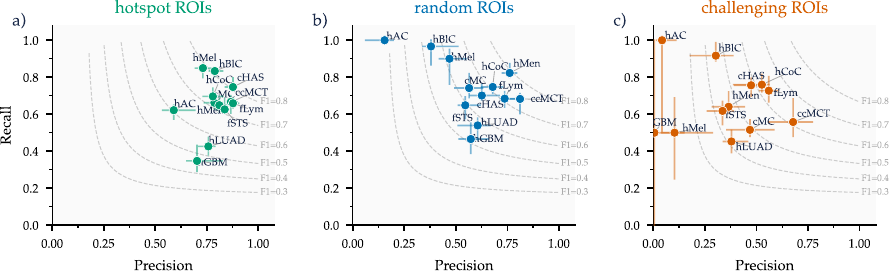}
    \caption{Precision and recall for all tumor domains and area types of Track 1 (mitotic figure detection). }
    \label{fig:precision-recall-areatypes}
\end{figure*}

\begin{figure*}
    \includegraphics[width=\linewidth]{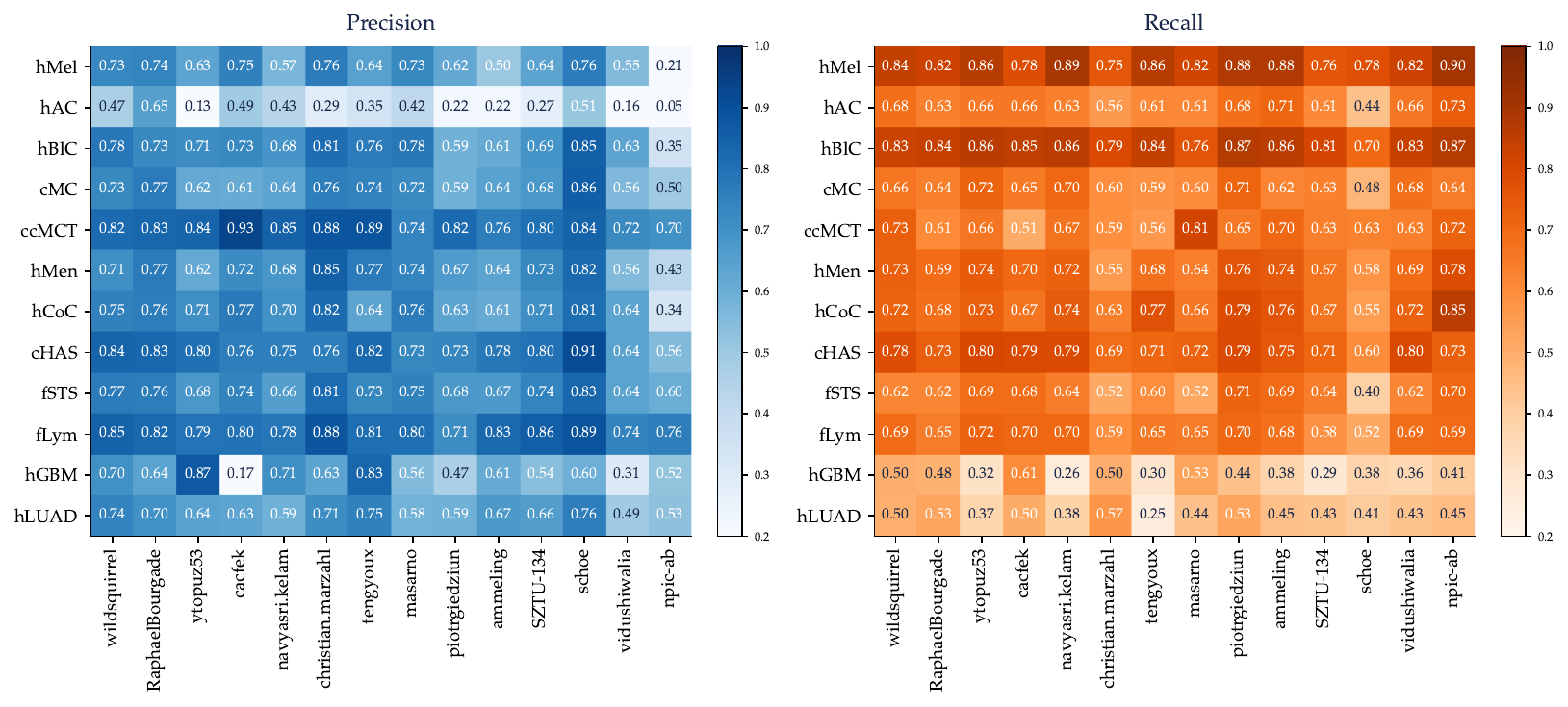}
    \caption{Precision and recall for all participants and tumor types in Track 1 (mitotic figure detection).}
    \label{fig:precision-recall-participants}
\end{figure*}

\begin{figure*}
    \includegraphics[width=\linewidth]{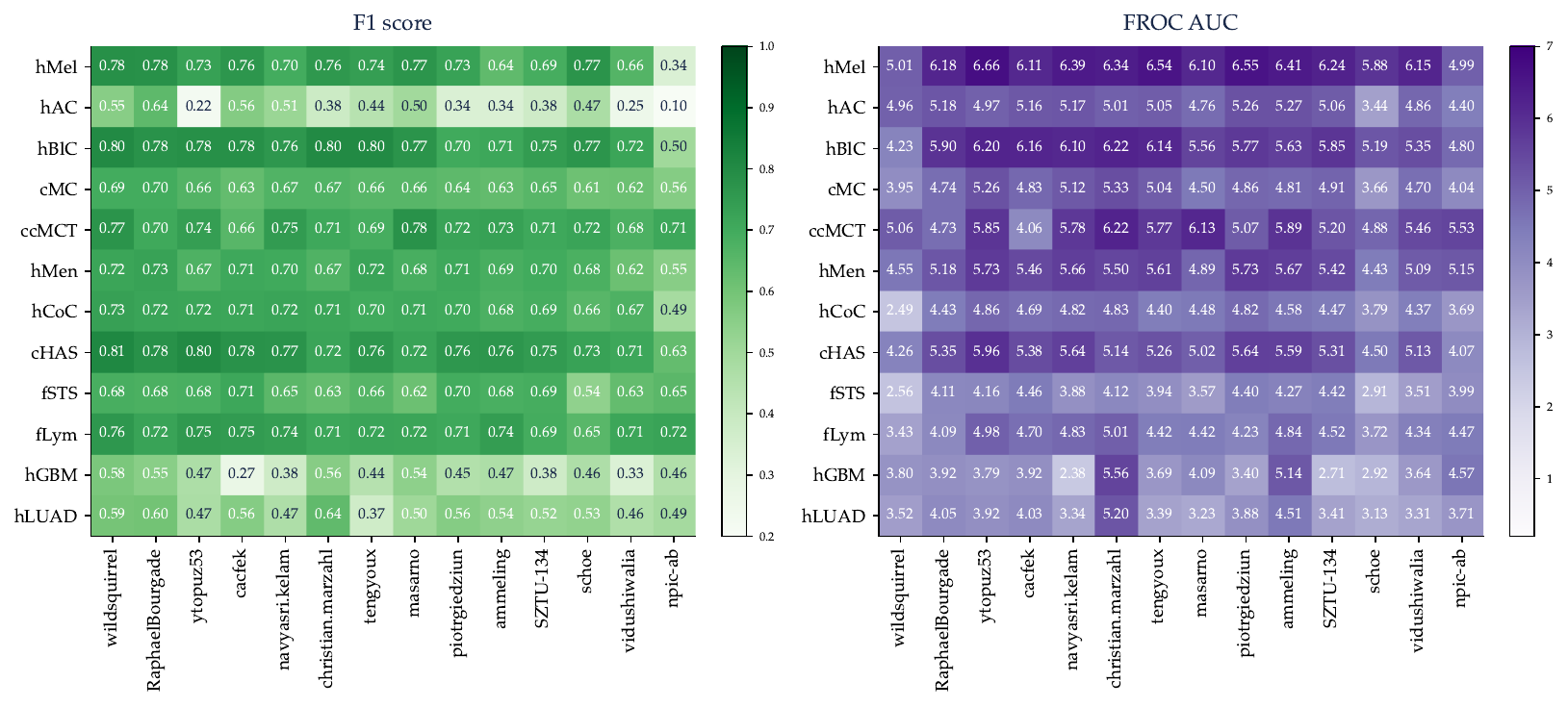}
    \caption{$F_1$ and FROC AUC for all participants and tumor types in Track 1 (mitotic figure detection).}
    \label{fig:f1-froc_auc-participants}
\end{figure*}

\begin{figure}
    \centering
    \includegraphics[width=\linewidth]{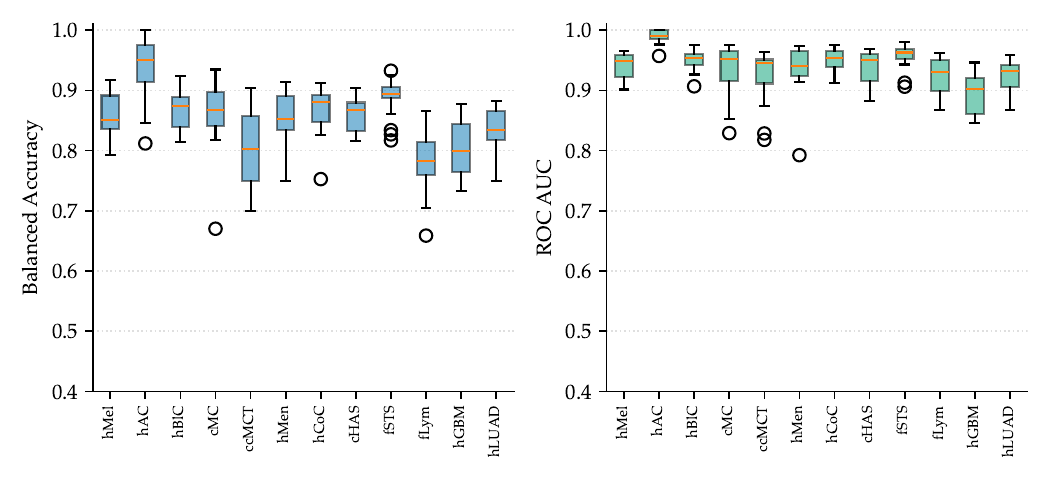}
    \caption{Balanced Accuracy (left) and ROC AUC (right) across tumor type of  challenge Track 2 (atypical classification), pooled across all participants.}
    \label{fig:ba_roc_auc_tumor_type_t2}
\end{figure}

\begin{figure}
    \centering
    \includegraphics[width=\linewidth]{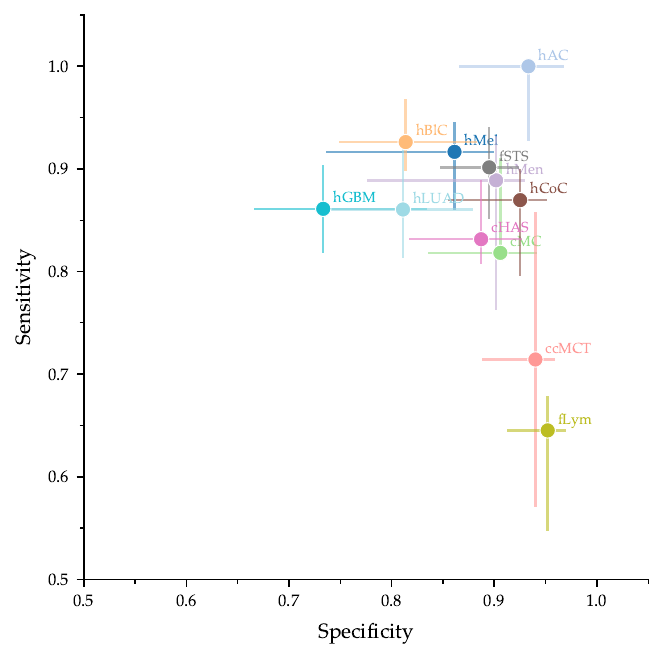}
    \caption{Specificity vs. Sensitivity by tumor type, pooled across all teams and samples, for Track 2 (atypical classification) of the challenge.}
    \label{fig:t2_ss_by_domain}
\end{figure}

\begin{figure*}
\includegraphics[width=\linewidth]{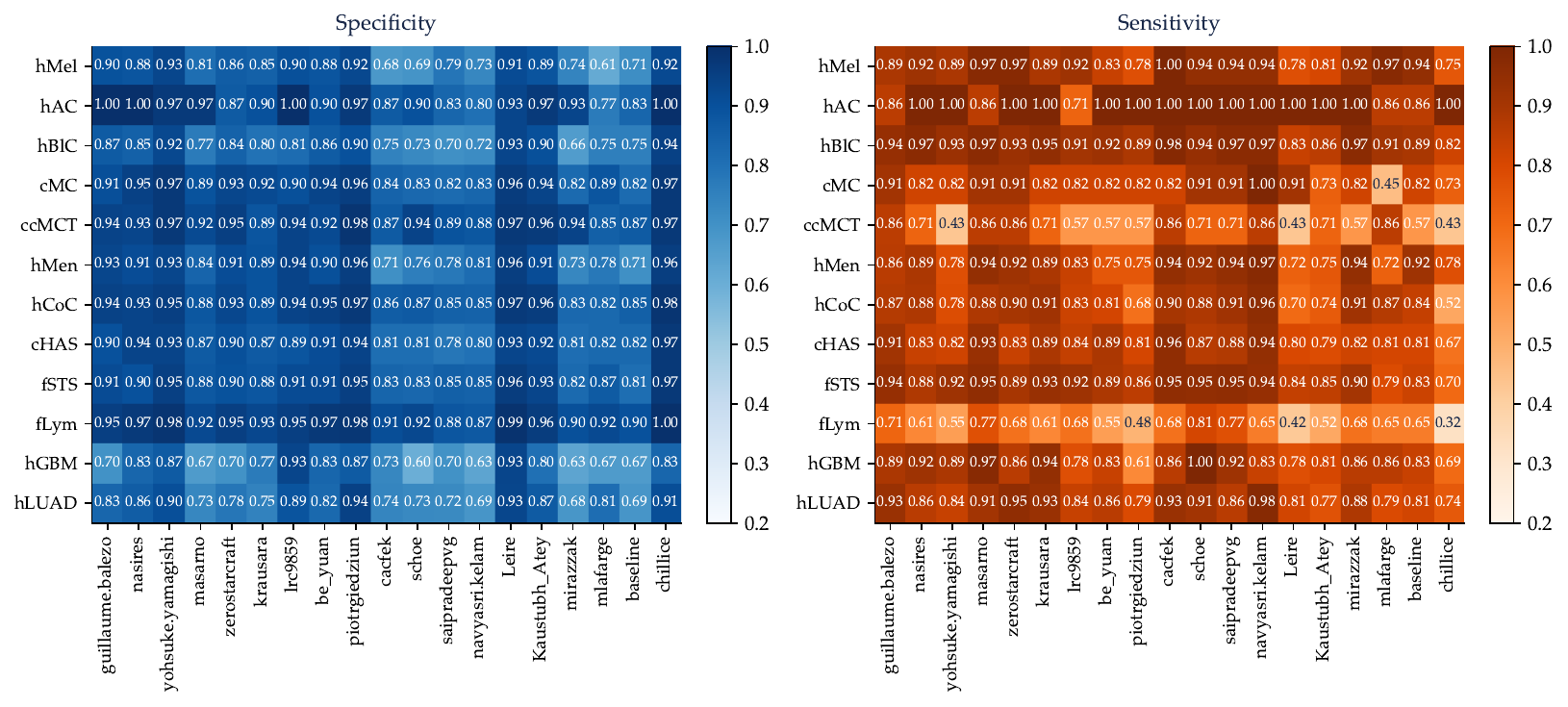}
    \caption{Specificity / Sensitivity per tumor type and team in Track 2 (atypical classification) of the challenge.}
    \label{fig:sens_spec_per_team}
\end{figure*}

\begin{figure*}
\includegraphics[width=\linewidth]{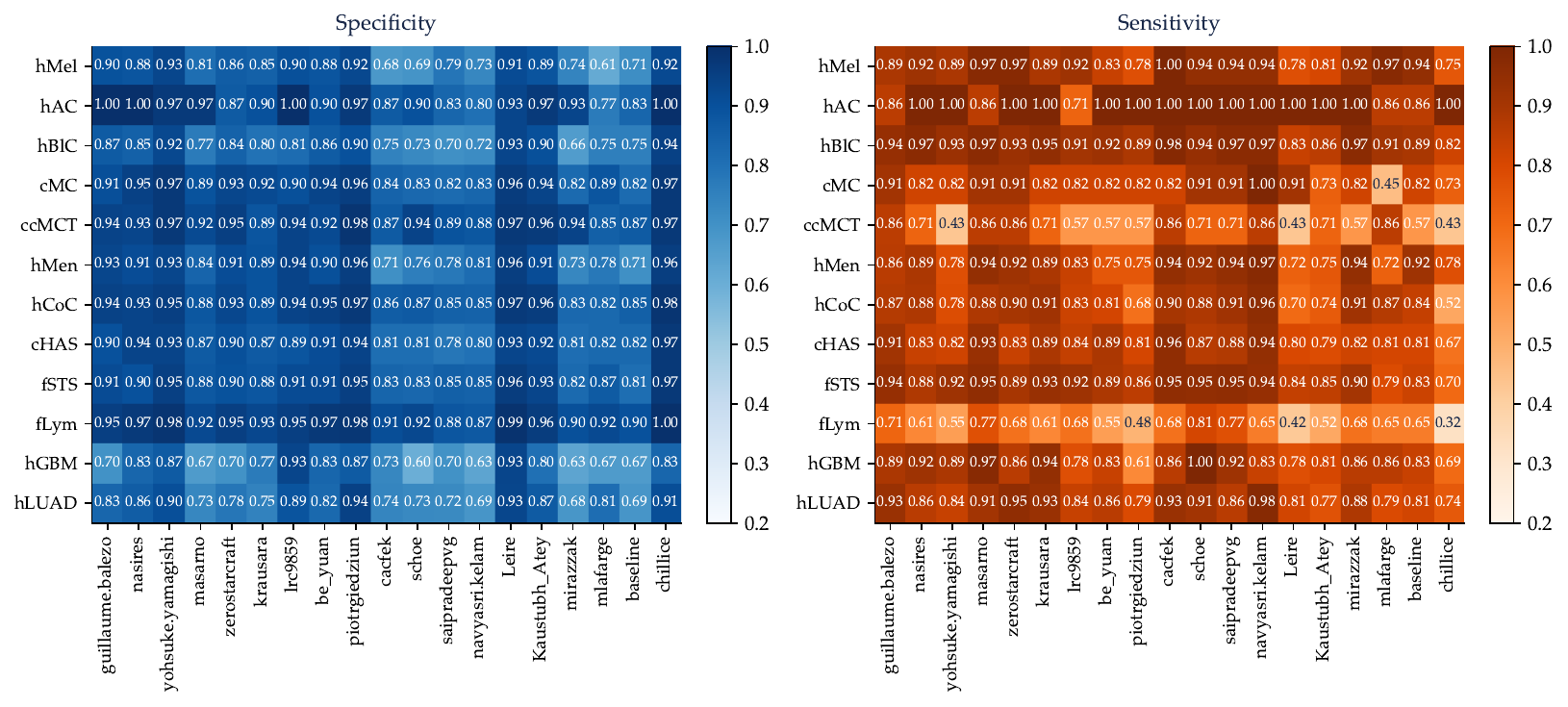}
    \caption{Balanced Accuracy and ROC AUC per tumor type and team in Track 2 (atypical classification) of the challenge.}
    \label{fig:BA_ROC_AUC_per_team}
\end{figure*}

\printcredits

%% Loading bibliography style file
%\bibliographystyle{model1-num-names}
\bibliographystyle{cas-model2-names}

% Loading bibliography database
\bibliography{cas-refs,proceedings}

%\vskip3pt

\end{document}